# Machine Learning and Citizen Science Approaches for Monitoring the Changing Environment

BY

SULONG ZHOU

A DISSERTATION SUBMITTED IN PARTIAL FULFILLMENT OF
THE REQUIREMENTS FOR THE DEGREE OF

DOCTOR OF PHILOSOPHY
(ENVIRONMENT AND RESOURCES)

AT THE
UNIVERSITY OF WISCONSIN - MADISON
2021

Date of final oral examination: 12/16/2020

The dissertation is approved by the following members of the Final Oral Committee:
    Janet Silbernagel, Professor, Planning and Landscape Architecture
    Song Gao, Assistant Professor, Geography
    Qunying Huang, Associate Professor, Geography
    Steve Ventura, Professor, Soil Science
    David Hart, Assistant Director, Sea Grant Institute

# Machine Learning and Citizen Science Approaches for Monitoring the Changing Environment

## Abstract

This dissertation will combine new tools and methodologies to answer pressing questions regarding inundation area and hurricane events in complex, heterogeneous changing environments. In addition to remote sensing approaches, citizen science and machine learning are both emerging fields that harness advancing technology to answer environmental management and disaster response questions.

Freshwater lakes supply a large amount of inland water resources to sustain local and regional developments. However, some lake systems depend upon great fluctuation in water surface area. Poyang lake, the largest freshwater lake in China, undergoes dramatic seasonal and interannual variations. Timely monitoring of Poyang lake surface provides essential information on variation of water occurrence for its ecosystem conservation. Application of histogram-based image segmentation in radar imagery has been widely used to detect water surface of lakes. Still, it is challenging to select the optimal threshold. Here, we analyze the advantages and disadvantages of a segmentation algorithm, the Otsu Method, from both mathematical and application perspectives. We implement the Otsu Method and provide reusable scripts to automatically select a threshold for surface water extraction using Sentinel-1 synthetic aperture radar (SAR) imagery on Google Earth Engine, a cloud-based platform that accelerates processing of Sentinel-1 data and auto-threshold computation. The optimal thresholds for each January from 2017 to 2020 are $-14.88$, $-16.93$, $-16.96$ and $-16.87$ respectively, and the overall accuracy achieves 92% after rectification. Furthermore, our study contributes to the update of temporal and spatial variation of Poyang lake, confirming that its surface water area fluctuated annually and tended to shrink both in the center and boundary of the lake on each January from 2017 to 2020.

Natural disasters cause significant damage, casualties and economical losses. Twitter has been used to support prompt disaster response and management because people tend to communicate and spread information on public social media platforms during disaster events. To retrieve real time situation awareness (SA) information from tweets, the most effective way to mine text is using Natural Language Processing (NLP). Among the advanced NLP models, the supervised approach can classify tweets into different categories to gain insight and leverage useful SA information from social media data. However, high performing supervised models require domain knowledge to specify categories and involve costly labeling tasks. This research proposes a guided Latent Dirichlet Allocation (LDA) workflow to investigate temporal latent topics from tweets during a recent disaster event, the 2020 Hurricane Laura. With integration of prior knowledge, a coherence model,

LDA topics visualization and validation from official reports, our guided approach reveals that most tweets contain several latent topics during the 10-day period of Hurricane Laura. This result indicates that state-of-the-art supervised models have not fully utilized tweet information because they only assign each tweet a single label. In contrast, our model can not only identify emerging topics during different disaster events but also provides multi-label references to the classification schema. In addition, our results can help to quickly identify and extract SA information to responders, stakeholders, and the general public so that they can adopt timely responsive strategies and wisely allocate resource during Hurricane events.

Citizen science and volunteered geographic information (VGI) are in high demand to encourage broader engagement of professional and unprofessional volunteers with their passions, knowledge, and experience. The prosperity of internet and mobile devices create new and increasing opportunities to sustain and maximize the contribution of spatial data from citizen science and VGI. More attentions have been paid on the quality and reliability of data collected from citizen science and VGI rather than the development process of applications and platforms for data collection. However, the questions on development are also dominant factors to affect the crowdsourcing data quality. For instance, what are the characteristics and functionalities of different user interfaces for VGI creation onsmart devices? How can stakeholders select relevant VGI for their specific tasks and needs? To addresses these questions, we introduced two pilot studies as the demonstration on how to select and develop mobile apps to collect field observations according to different research purposes. The case studies included detailed and valuable insights in the comparison metric of 3-tier architecture from both user and developer perspective. Additionally, we also shared the complete process to customize popular participatory mobile applications, Surver123 and AppStudio, provided recommendations for different stakeholders, and found that having clear requirements and being familiar with functionalities of the app development tools were two prominent features to develop and improve customized applications.

# Contents

# Listing of figures

# Acknowledgments

Firstly, I would like to express my sincere gratitude to my advisor Prof. Silbernagel for the continuous support of my Ph.D study and related research, for her patience, motivation, and immense knowledge. Her guidance helped me in all the time of research and writing of this thesis. I could not have imagined having a better advisor and mentor for my Ph.D study.

Besides my advisor, I would like to thank the rest of my thesis committee: Prof. Huang, Prof. Gao, Prof. Ventura and Dr. Hart, for their insightful comments and encouragement, but also for the hard question which encouraged me to widen my research from various perspectives.

My sincere thanks also goes to my dear friend, Pengyu Kan, we spent countless days and nights working on brainstorming, processing data, tuning models and debugging during COVID-19. I hope he can receive as many graduate school offers as possible.

For work in Chapter 1, I appreciate the award of Incubator program of Institute for Regional and International Studies at University of Wisconsin-Madison to support our research and the collaboration with Annemarie Schneider of SAGE (Sustainability and the Global Environment) at the University of Wisconsin-Madison. I also gratefully acknowledge Poyang lake National Nature Reserve staff who collaborated with us to collect ground truth data.

For work in Chapter 2 and 3, I appreciate the award from Dr. Hart and the Lake Superior National Estuarine Research Reserve to support my travel to ESRI and learn AppStudio for participatory mobile application development.

Finally, I would like to thank the Department of Geography and the Department of Computer Sciences to offer me continuous scholarship to make me survive along the PhD paths.

# 0

# Introduction

Earth is experiencing dynamic changes. Some are periodical as water extent of inland freshwater lakes shrinks, and some are prone to cause large natural disaster events such as hurricanes, one of the largest environmental drivers of changes in coastal systems. These changes are strongly associated with their affiliated ecosystems, regional economies, and the lives and security of residents. Retrieving relevant and timely information on these changes is crucial to maintain sustainable development and response to sudden disasters.

To monitor and track seasonally or randomly altered environments and acquire timely information, Remote Sensing (RS) and Geography Information System (GIS) have been widely considered as powerful data collection, processing and analysis technology. At the same time, with the growth of social media platforms and citizen science opportunities have expanded steadily over last decades, supporting agency research, and monitoring, and providing valuable education and environmental awareness. As affordable technologies have been widely employed, the amount of remote sensing and crowdsourced data collection has scaled up subsequently. Consequently, artificial intelligence and machine learning have been in high demand to filter, process and analyze those "big data". As the result of explosion of data size and volume, the models and platforms have emerged one after another. It is essential to select the best suitable models and platforms for a research. However, some individuals only pursue the most advanced and complicated models and platforms which is inefficient and even harmful.

To generalize the procedure to select and utilize suitable data, models and platforms for environmental studies, my dissertation will propose an abstracted $\pi$ framework (upper part of Figure 1) to provide an insight of integration of machine learning, citizen sciences and environmental sciences. The top bar represents the width of applications of environmental sciences while the two legs represent two different pipelines towards the applications. The pipeline consists of three components: data, platform and models. The A and B differentiate the data source either from authoritative data or crowdsourced data. The authoritative data refers to institutional and government data, such as administrative boundary map, satellite imagery collections, and official Digital Elevation Models (DEMs) data. These GIS/RS data are well-structured and quality assured but sometimes controlled under access policies and digital rights. On the other hand, the crowdsourced data are unstructured and uncalibrated citizen science data collected across social medias from ubiquitous mobile devices and web services. The social media platforms, Twitter, YouTube, and Flickr, provide millions of audios, videos, photos and texts. To process the high volume data and support scale-up research collaboration, cloud platforms (Google Earth Engine, Google Colab, GitHub, ArcGIS Online) have emerged to replace the traditional stand-alone platforms. An additional benefit of using cloud platforms comes from their shareable code base and reusable models. In terms of models, different supervised and unsupervised models have been applied to process both authoritative data and crowdsourced data. Examples include using tree models, such as decision tree and random forest, to classify remote sensing imagery and using clustering and neural network models, such as Latent Dirichlet allocation, convolutional neural network, and recurrent neural network to process texts and messages.

By presenting international case studies, my dissertation will follow the proposed $\pi$ framework to examine how remote sensing/GIS, machine learning, and citizen science perform in monitoring changing environment. The overall structure of this dissertation is shown in the lower part of Figure 1, Chapter 1 focuses on an implementation of my $\pi$ framework using pipeline A for land cover research while Chapter 2 focuses on the other implementation of my $\pi$ framework for retrieving and verifying situational awareness. In addition, Chapter 3 demonstrates how in-house crowdsourced data collection application can supplement previous Chapters.

Chapter 1 * will center on an international case study of remote sensing in surface water extent inventory and monitoring at Poyang Lake. Poyang Lake is the largest freshwater lake in China with extremely variable and altered hydrology. Since the hydrological dynamics includes seasonal and annual variation of inundated areas in Poyang Lake, up-to-date water surface extent was required. This case study employed Sentinel-1 satellite imagery data and Google Earth Engine to detect the seasonal water change of Poyang Lake.

Hurricane events have been regarded as one of the most damaging natural disasters. Using social media to assist disaster response and management has steadily grown in the last decade and has made substantial contributions. How to retrieve timely information with the help of advanced machine learning models, such as Natural Language Processing, has attracted much attention. In Chapter 2 [†], a guided Latent Dirichlet Allocation (LDA) framework that integrates expert domain knowledge, coherence model, topics visualization and validation from official report for detecting and analyzing latent topics during hurricane events will be proposed.

---

*This chapter has been published in the ISPRS International Journal of Geo-Information (ISSN 2220-9964) in the year of 2020[136]

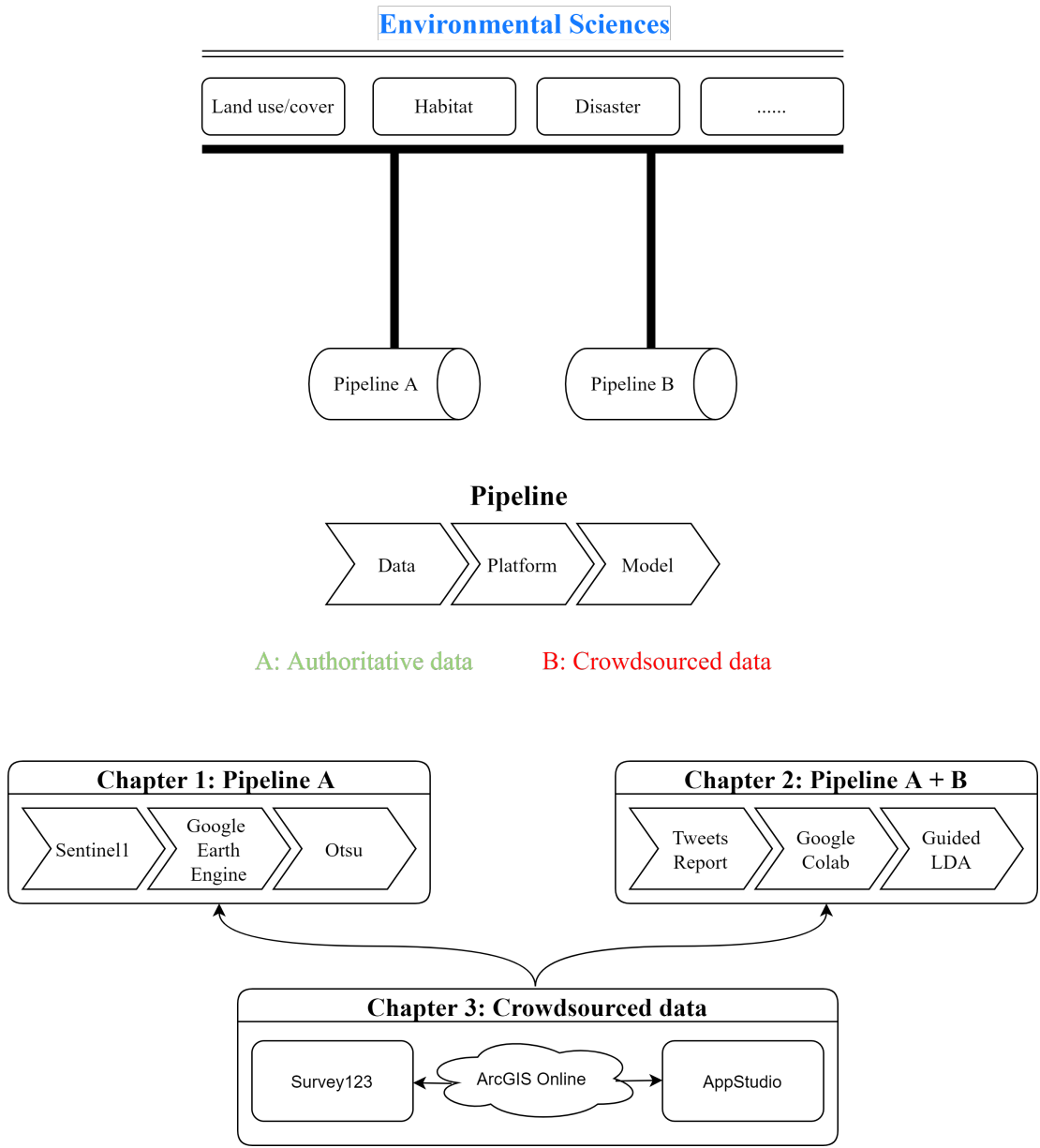

**Figure 1:** Abstracted $\pi$ framework and dissertation structure.

Citizen science and volunteered geographic information (VGI) are in high demand to encourage broader engagement of not only scientists but also a variety of participants including government agencies, organizations, and public. The prosperity of internet and mobile devices create new and increasing opportunities to sustain and maximize the contribution of citizen science and VGI. Spurred by the advanced geographic-enabled applications such as Survey123 and App Studio developed by Esri, Chapter 3 will introduce two pilot studies as the demonstration on how to develop mobile apps to collect field observations according to different research purposes, and discuss the learned lessons and potential future development.

Environmental science is an interdisciplinary study of how natural and man-made processes interact with one another and ultimately affect the various biomes of Earth that combines the fields of ecology, biology, hydrology, geography, social science and more. Nowadays, computer science has created productive tools and models for varieties of disciplines in both theoretical and practical processes. My PhD pathway aims to undertake the application of advanced machine learning models and software development suites in environmental studies. I plan to learn and verify how efficiently and effectively these technologies can help to facilitate data acquisition and mining to monitor the changing environment. In return, I want to write this dissertation to demonstrate how different environmental studies can benefit from computer science regarding three principle components, data, platform and model in my proposed $\pi$ framework. The major contributions include strengthening the link between environmental studies and computer science, showcasing successful international

---

[†]This chapter has been published in the Journal of Information Science in the year of 2021 [135]

case studies, recommending development process of in-house citizen science application, and offering reusable and reproducible models across cloud platforms. In this case, my dissertation can be a pebble in the river of prosperity of integration of computer science and environmental science.

**1**

# Application of Image Segmentation in Surface Water Extraction of Freshwater Lakes using Radar Data

## 1.1 INTRODUCTION

Water is significant for all ecosystems on Earth. The presence of surface water on Earth mainly consists of oceans, lakes and rivers[90]. The extent of lakes accounts for nearly 3% of the surface[29] and is endowed with irreplaceable functions to supply water[11], control flooding[107], sustain species[52] and provide ecosystem services to nations and regions[94] due to the unique role of water in climate[50], biological diversity[39] and human wellbeing[24]. Meanwhile, natural phenomena and human activities affect the variation of water occurrence in response, especially the water dynamics of inland freshwater lakes[121]. Timely monitoring of freshwater lake surface is indispensable for sustainable development[118] and regional and global ecosystem dynamics[48].

Remote sensing, the science and art of detecting objects from a distance, has been the most common approach to monitor and analyze land features for several decades[20]. In imagery, land features are typically represented as mixed classes of different vegetation cover and surface types. There are many satellite-based sensors that differ in terms of temporal and spatial resolution, corresponding to revisit time and ground area represented by a pixel, respectively. Medium resolution imagery is the most widely used for lake water surface detection, (with approximately 10 days revisit time and each pixel ranging from 10 to 30 m), due to its open access compared to the cost of acquiring higher resolution imagery[111] and are less prone to the mixed pixels problem of coarse resolution imagery[19]. Aside from temporal and spatial resolution, there are both passive sensors and active sensors. Passive sensors, known as optical systems, have been employed since the 1970s when the first satellite sensor, Landsat multispectral scanner (MSS), was launched into space[126]. However, lack of vertical information, issues with wetland vegetation overlapping canopy, and haze and cloud cover problems have largely impeded the accuracy of results[62]. Thus active sensors, particularly radar systems, have also contributed to remote sensing of water dominated systems, such as lakes. Radar backscatter is sensitive to moisture content and roughness of landscape, and the wavelength of C-band Sentinel-1 sensor enables penetration of both clouds and thick canopies to deal with the challenges of complicated weather and flora conditions[14].

Nevertheless, the procedure of processing Sentinel-1 radar data involving data acquisition, calibration, speckle filtering, geometry and terrain correction, classification and validation[116] is extremely time consuming with use of traditional image process platforms, even those with built-in toolboxes, such as ENVI and ERDAS software packages[131]. This cost can limit the timeliness and efficiency of research. With the help of high-performance computing and network systems, Google Earth Engine (GEE) allows online processing and analysis of radar imagery by writing light-weight scripts with a Google account, speeding the process in a cloud-based platform[43]. The plethora of data catalogs and innovative processing algorithms provided by GEE can effectively eliminate the barriers caused by the traditional platforms. The water detection algorithms based on radar sensors have emerged in several categories: thresholding, classification and object-based image analysis. In general, thresholding has commonly been adopted to discriminate water from nonwater surface in the logarithmic representation of the radar imagery, where the water and nonwater features are shown as two Gaussian distributions in the histogram of backscatter coefficient of radar data in dB scale. Although it is limited by double bounce scattering issues because waters beneath vegetation layers may cause extra radar backscatter [100], thresholding is still an efficient and simple method for water extraction of rural areas in winter season with less complicated vegetation coverage.

One classical method to select the threshold is to manually pick the smallest valley values between the two peaks of distributions based upon visual inspection by the researcher. The main issue of this method is the bias caused by each individual observer. The solution to offset the researcher's observation bias is to apply computer programming to select a less biased lowest point in the valley, which can be computationally efficient in linear time. However, the intensity histogram presented by radar imaging may not necessarily provide a sharp valley but usually a flat region between the peaks. Thus, it will be less accurate or reasonable to pick the smallest valley value in this case, as the value of the selected point may deviate slightly from the value of its neighboring points in the open intervals next to the selected point. Furthermore, due to the noise in radar detection, the strict convex property is not guaranteed in the valley region between the two peaks. In other words, there may exist multiple local peaks and minimums which are close to each other. In this case, the method of picking the smallest valley value is badly influenced by the noise.

The Gaussian Mixture Model is another conventional method for binary classification based on distribution. The distribution of water and nonwater objects in the radar intensity (dB) histogram presents approximately as two Gaussian Distributions with separate means $\mu_1$ and $\mu_2$ and standard deviations $\sigma_1$ and $\sigma_2$ [127]. One of the distributions is the conditional probability of the dB value of the water pixels while the other is the conditional probability of the dB value of the nonwater pixels. The objective of this model is to maximize conditional probability of the prediction $\hat{y}$ given any dB values ($x$). According to the Bayes Theorem, this equates to maximizing the multiplication of the conditional probability of $x$ over $\hat{y}$ and the marginal probability of $\hat{y}$.

However, the issue with such formulation of the problem is based on the assumption of the prior distribution of water and land as a Gaussian Distribution. However, such an assumption cannot be directly assumed to be correct for universal cases. Moreover, the distribution parameters $\mu_1, \sigma_1, \mu_2,$ $\sigma_2$ are unknowns. The researcher also needs to identify estimators for these four parameters through the density diagram. Possible solutions for estimation of these unknown parameters can be iterative methods such as Expectation Maximization Method[125], however, it is unstable for two reasons. First, the iteration process is time consuming to reach a satisfied accuracy. Second, it is also likely to be constrained in some local optimum points and thus never reaches global optimal solution[86].

Instead, we propose to use the Otsu Method to solve this thresholding problem. The Otsu Method is an unsupervised method and it was initially designed to select a threshold to separate an object out of its background, through the gray-level histogram of the image[86]. In application, the Otsu Method can be widely extended to work on other density histograms or distributions other than gray-level histogram from images and can also be applied for multi-thresholding problems. The Otsu Method is a better approach for this problem as compared to some conventional solutions because it automatically selects a threshold from two mixed distributions through the density histogram[86]. In addition, the Otsu Method does not require prior knowledge nor assumptions of the distribution of objects[86]. Furthermore, the Otsu Method is equivalent to the K-Means Method but the Otsu Method can provide the global optimal solution, while K - Means Method may be limited to the local optimum point[117]. Although it is computationally complex and heavy because of iterative searching[117], GEE can speed up the Otsu Method with its cloud computing platform. For instance, the Otsu Method has been applied on the cloud-free Landsat TM images for urban land cover detection, which focused on differentiating the urban land and nonurban land region in Haidian District of Beijing, China[4]. This research resulted in an accuracy of 84.83% for the Otsu Method, which was larger than the accuracy of 74% for the conventional postclassification change detection method[4]. Another study used the Otsu Method on the SAR data for the detection of oil spills over sea surfaces, which tried to find a threshold on the radar data to draw the edge of spilled oil film floating over the sea[132]. It examined the Penglai oil field and the Gulf of Dalian, resulting in an error rate of 3.0% on the Penglai oil field and an error rate of 13.0% on the Gulf of Dalian for the Otsu Method[132]. Even though the Otsu Method has already been widely applied in thresholding problems, it has been seldom used for surface water extraction. Furthermore, most previous studies do not provide algorithms and detailed scripts for implementation of the Otsu Method. Thus, we were interested in the application of the Otsu Method for surface water detection and providing reusable code for future implementation.

Therefore, the objectives of the present work are to:

1. Implement the Otsu Method and write reusable scripts to automatically select thresholds for surface water extraction using Sentinel-1 data on Google Earth Engine

2. Analyze the advantages and disadvantages of an unsupervised classifier from both mathematical and application perspectives

3. Contribute to the knowledge base of hydrological variation at Poyang lake by mapping surface water extent of the lake in January 2017, 2018, 2019 and 2020

## 1.2   Materials and Methods

### 1.2.1   Study Area

Poyang lake, the largest freshwater lake in China, is located between Nanchang City and Jiujiang City, to the north of Jiangxi Province. The basin crosses from 28°22′ to 29°45′ N and 115°47′ to 116°45′ E, which belongs to a humid, subtropical monsoon climate zone, with an average annual temperature of 17.5 °C and average annual precipitation of 1665 mm[129]. Poyang lake basin is fed by the Xiu, Gan, Fu, Xin and Rao rivers, while the basin connects to the Yangtze river through an outflow channel at the north end of the lake (Figure 1.1). The lake has a surface area of approximately 4000 square kilometers at its summer high-water level[102,18]. Beyond its size, Poyang lake is also significant for several economic and ecological reasons. For instance, Poyang lake's aquatic ecosystems are wintering home to thousands of migratory waterbirds, including the Siberian crane—a critically endangered species whose 4000 surviving individuals spend their winters almost solely in the wetlands around Poyang lake[75]. However, Poyang lake has undergone a series of significant transformations that threaten the variability and critical habitats in the region. While surface water areas have traditionally fluctuated on a seasonal scale—peaking in the summer and receding in the winter—large interannual declines in mean water level have been observed in recent years[130]. The substantial variations of the surface water area and dramatic seasonal water level fluctuations of 8 to 22 m each year are caused by the regional hydrological regime, which is controlled both by the five catchment rivers and the Yangtze River[38]. Additionally, groundwater dynamics are highly affected by the variations in the lake water level, rather than local precipitation, indicating a close hydraulic relationship between groundwater and the lake[68].

### 1.2.2  Platform and Data

Google Earth Engine (GEE, `https://earthengine.google.com`) consists of a multipetabyte satellite imagery data catalog colocated with a high-performance, intrinsically parallel cloud computation service. Users can access GEE through an Internet-accessible application programming interface (API) and an associated web-based interactive development environment (IDE) that enables rapid prototyping and visualization of results. This cloud computing platform not only makes it easy to access most of the geospatial datasets but also enables high throughput analysis. There are many examples where environmental scientists empowered their research with help of GEE, such as population mapping[88], cropland mapping[103], extraction of glacial lakes[21] and probabilistic wetland mapping[49].

Sentinel-1 is the first Copernicus Program satellite constellation deployed by the European Space Agency. This space mission is composed of two satellites, Sentinel-1A and Sentinel-1B, carrying a C-band synthetic-aperture radar instrument which collects data in all weather, day or night[115].

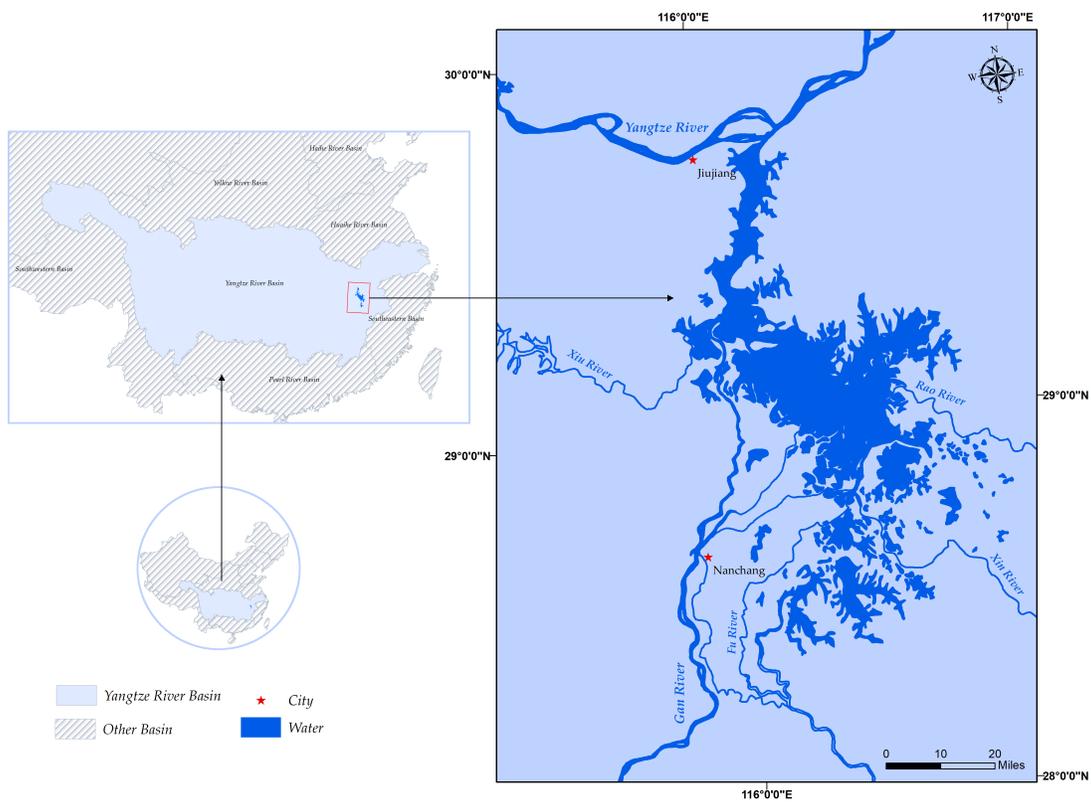

**Figure 1.1:** Location of Poyang lake within the Yangtze River Basin (**left**) and at an average level (**right**).

Since radar sensors have the advantage in detecting moisture and water because of their ability to penetrate clouds, Sentinel-1 is one of the most common datasets for surface water detection[91] and flood mapping[16].

The winter low-water season of Poyang lake provides important foraging habitat and wintering area for many waterbirds of special concern, including the critically endangered Siberian crane. Because of the importance of water level during this time, we looked at images taken in January over subsequent years. We loaded Sentinel-1 Level-1 IW GRD images from the data catalog of GEE from January 2017–2020 (Table 1.1). The imagery acquired on January of 2020 was used to evaluate our Otsu Method implementation on GEE. The others were used to analyze the water area change in January across 4 years from 2017 to 2020.

**Table 1.1:** Data Collection for Inundation Area Detection.

| Platform | Type | Spatial Resolution | Date | Band | Instrument | Orbit |
|---|---|---|---|---|---|---|
| Sentinel 1 B | GRD | 10 m | 01/04/2020 | VV | IW | 19666 |
| Sentinel 1 B | GRD | 10 m | 01/09/2019 | VV | IW | 14416 |
| Sentinel 1 B | GRD | 10 m | 01/02/2018 | VV | IW | 8991 |
| Sentinel 1 B | GRD | 10 m | 01/07/2017 | VV | IW | 3741 |

### 1.2.3 Otsu Method

In this section, we firstly introduce the main idea of the Otsu Method[86,117] in a general framework and then we discuss how the Otsu Method is applied on this thresholding problem with a radar value density histogram.

Here we use the following notations:

- set $C$ is the index set of all elements.

- $i$ is the index of $i$-th element, where $i \in C$ refers to $i$-th element belonging to the whole set $C$

we are considering. $x_i$ is the value for this $i$-th element and without loss of generality, we can assume that $x_i$ are sorted. Explicitly, $x_i < x_j$, if $i < j$, for $i, j \in C$.

- $p_i$ is the probability or density of the element $i$. It is clear that $\sum\limits_{i \in C} p_i = 1$.

- we try to split up the set $C$ into two disjoint subclusters of index $C_0, C_1$, where $C_0, C_1 \subset C$ and $C_0 \dot{\bigcup} C_1 = C$.

- $\mu_j$ is the center or the mean value of cluster $C_j$, for $j = 0, 1$:

$$\mu_j = \frac{\sum\limits_{i \in C_j} p_i \cdot x_i}{\sum\limits_{i \in C_j} p_i} \quad , \text{for } j = 0, 1$$

- $\mu$ is the center or the mean value of the whole set $C$:

$$\mu = \sum_{i \in C} p_i \cdot x_i = \sum_{i \in C_0 \dot{\bigcup} C_1} p_i \cdot x_i$$

- $V_j$ is denoted as the inner-variance of the cluster $C_j$, which is defined as the weighted summation of the squared distance of cluster $C_j$'s each data point from its center $\mu_j$, for $j = 0, 1$:

$$V_j = \sum_{i \in C_j} p_i \cdot (x_i - \mu_j)^2 \quad , \text{for } j = 0, 1$$

- $V_{0,1}$ is denoted as the interclass variance between the cluster $C_0$ and cluster $C_1$ [86], which is defined as the weighted summation of the squared distance of each cluster's center $\mu_j$ from the center of the whole set $\mu$:

$$V_{0,1} = \sum_{j=0,1} ((\sum_{i \in C_j} p_i) \cdot (\mu_j - \mu)^2)$$

- $V$ is denoted as the total-variance, which is defined as the weighted summation of the squared distance of all data points from the center of the whole set $\mu$. Furthermore, we can see that $V$ is actually exactly the variance $\sigma_C^2$ of the set $C$:

$$V = \sum_{i \in C} p_i \cdot (x_i - \mu)^2$$

The main idea of the Otsu Method is to minimize the summation of the inner-variance $V_j$ of all clusters $C_j$, which is called intraclass variance[86]. The inner-variance of a cluster shows the summation of squared distance of each element to the center of the cluster as we defined, and the smaller value of the inner-variance presents the closer distance of each point toward the center of the cluster, which shows a closer relationship or higher similarity that the elements in this cluster share. Therefore, the best separation of the whole set of elements should group the similar elements in the same cluster as optimally as possible. In mathematics, this is equivalent to minimizing the summation of inner-variance inside each cluster. The objective function is formulated as follows:

$$\min_{C_0, C_1} \sum_{j=0,1} V_j = \min_{C_0, C_1} \sum_{j=0,1} \sum_{i \in C_j} p_i \cdot (x_i - \mu_j)^2 \tag{1.1}$$

Furthermore, the summation of each cluster's inner-variance and the interclass variance should be equal to the total-variance of the whole set[86], which is a constant for a fixed data set.

$$V = \sum_{j=0,1} V_j + V_{0,1} \tag{1.2}$$

Therefore, the previous objective function Equation (1.1) is equivalent to maximizing the interclass variance $V_{0,1}$:

$$\max_{C_0, C_1} \sum_{j=0,1} ((\sum_{i \in C_j} p_i) \cdot (\mu_j - \mu)^2) \tag{1.3}$$

Now, in applying the Otsu Method on the density histogram, we can have:

- The set of all possible bin's values on the density histogram as $\Theta$, which is also the hypothesis space for the estimation of the threshold.

- The density corresponding to the bin with value $\theta$ is denoted as $p_\theta$ and we should have $1 = \sum_{\theta \in \Theta} p_\theta$.

- For each bin value $\theta \in \Theta$, we can put a corresponding index $i_\theta$ into the indexed set $C$, where $i_\theta$-th bin on the density histogram has a bin value of $\theta$. Therefore, $x_{i_\theta} = \theta$ and $p_{i_\theta}$ is equal to the density $p_\theta$ of the bin with value $\theta$ from the density histogram.

- The final prediction of the threshold is denoted as $\hat{\theta} \in \Theta$, which corresponds to the index $i_{\hat{\theta}} \in C$.

- the two separate clusters formed by a threshold $\theta$ is $C_0 = \{i : i < i_\theta, i \in C\}$ and $C_1 = \{i : i \geq i_\theta, i \in C\}$.

The final estimation of the threshold $\hat{\theta}$ should be the one based on which the subcluster $C_0^*$ and $C_1^*$ can reach the optimal value of the objective function in Equation (1.3). Then, we can have:

$$i_{\hat{\theta}} = \min_{i \in C_1^*} i \tag{1.4}$$

$$\hat{\theta} = x_{i_{\hat{\theta}}} \tag{1.5}$$

This optimization problem can be solved in at most quadratic polynomial time of the size of the finite set $\Theta$, i.e., the time complexity will be $O(|\Theta|^2)$. One possible implementation as shown in Algorithm 1 is to iterate through the finite set and record the element in the set that provides

the highest value for the objective function. Each inner iteration takes linear time to calculate the objective function.

---

**Algorithm 1:** OtsuMethodFindOptimalThresholding (Time: $O(|\Theta|^2)$)

---

**Input:** $\{(\theta_i, p_{\theta_i})\}_{i=1}^{|\Theta|}$ is the set of bins for the density histogram.
**Output:** $\hat{\theta}$ is the final prediction of the optimal threshold between the two classes.
**Procedure:**
$\hat{\theta} \leftarrow \infty$
*objective_value* $\leftarrow 0$
**for** *each bin i in the input density histogram* **do**
  $\quad$ $C_0 \leftarrow \emptyset$
  $\quad$ $C_1 \leftarrow \emptyset$
  $\quad$ **for** *each bin j in the input density histogram* **do**
    $\quad\quad$ **if** $\theta_j < \theta_i$ **then**
      $\quad\quad\quad$ $C_0$ adds $j$
    $\quad\quad$ **else**
      $\quad\quad\quad$ $C_1$ adds $j$
    $\quad\quad$ **end**
  $\quad$ **end**
  $\quad$ *curr* $\leftarrow$ Compute objective function value based on Equation (1.3) with using $C_0, C_1$
  $\quad$ **if** *curr > objective_value* **then**
    $\quad\quad$ $\hat{\theta} \leftarrow \theta_i$
  $\quad$ **else**
  $\quad$ **end**
**end**
**return** $\hat{\theta}$

---

We can further improve the time complexity of the Otsu Method into linear time complexity of $O(|\Theta|)$. If we store the value of $\mu_0$ and $\mu_1$ from previous outer loop iteration, then it will take constant time $O(1)$ for recomputing the objective function value based on the Equation (1.3) for the newly updated $C_0$ and $C_1$ in this current round.

Because the Otsu Method iterates through all the possible values for the threshold and compares the objective values with all these possible thresholds, the implementation of Otsu Method in Algorithm 1 provides a global optimal solution for the objective function in Equation (1.3).

### 1.2.4 WATER DETECTION

Once the preprocessing procedure was completed by GEE, the histogram of VV band was generated, and the Otsu method was used to search over the thresholds that are represented by the bins in the histogram. The optimal threshold was computed to classify the data, where the partition whose values are smaller than the threshold are labeled as water while the partition whose values are larger than the threshold are labeled as nonwater. In order to reduce the effect of double bounce scattering issues, we defined the label of water as purely open water area, while the label of nonwater included the submerged and emergent aquatic vegetation and land features. The specific implementation of the Otsu Method can be found through the link: source code for the Otsu Method (by Sulong Zhou). The postprocessing procedure that removes noise and improves the quality of the classified output involving mask extraction, majority filtering and boundary clean was carried out in ArcGIS to remove the water bodies not geographically related to Poyang lake, small islands of pixels and odd edge of clusters.

### 1.2.5 ACCURACY ASSESSMENT

We denote a point $x$ with its true label $y$ drawn from the true distribution $\mathcal{D}$ as $(x, y) \sim \mathcal{D}$, where the true distribution $\mathcal{D}$ is actually unknown. Specifically, $x$ is the radar dB value for a pixel and defined as:

$$y = \begin{cases} 1 & \text{, if the true label of } x \text{ is Water} \\ 0 & \text{, if the true label of } x \text{ is Non-Water} \end{cases}$$

Based on the estimation of the optimal threshold $\hat{\theta}$ from the Otsu Method, we can provide the prediction of the label $\hat{y}$ for point $x$ as:

$$\hat{y} = f_{\hat{\theta}}(x) = 1_{x < \hat{\theta}} \tag{1.6}$$

where,

$$1_{x < \hat{\theta}} = \begin{cases} 1 & \text{, if } x < \hat{\theta} \\ 0 & \text{, otherwise} \end{cases}$$

*Accuracy* measures the agreement between a standard assumed to be correct and a classified image of unknown quality [108]. Classification errors occur when a pixel (or feature) belonging to one category is assigned to another category. Errors of omission occur when a feature is left out of the category being evaluated; errors of commission occur when a feature is incorrectly included in the category being evaluated [36]. An error of omission in one category will be counted as an error in commission in another category. Explicitly, for a pixel's dB value $x$ and its true label $y$ from the unknown true distribution $\mathcal{D}$, the error happens when $\hat{y} \neq y$. Therefore, accuracy can be mathematically defined as the follow:

$$\text{Accuracy} = \mathbb{E}_{(x,y) \sim \mathcal{D}} 1_{\hat{y} = y} \tag{1.7}$$

Since the true distribution $\mathcal{D}$ is unknown, it is not possible to calculate the accuracy through Equation (1.7). Therefore, we need an estimator to estimate such accuracy. One possible way to estimate is based on the Empirical Distribution [42]. A test set $T = \{(x_i, y_i)\}_{i=1}^n$, which forms an empirical distribution $\hat{D}$, is used to approximate the true distribution $\mathcal{D}$, where each element $(x_i, y_i)$ is independently and identically (*i.i.d*) drawn from the true distribution $\mathcal{D}$.

$$\{(x_i, y_i)\}_{i=1}^n \overset{i.i.d}{\sim} \mathcal{D}$$

The empirical estimator of accuracy $\widehat{Accuracy}$ for classification function $f_{\hat{\theta}}$ can be expressed as:

$$\widehat{Accuracy} = \frac{1}{n} \cdot \sum_{(x_i, y_i) \in T} 1_{\hat{y}_i = y_i} \quad , \text{ where } \hat{y}_i = f_{\hat{\theta}}(x_i) \tag{1.8}$$

Since $(x_i, y_i) \in T$, for $\forall i = 1, \ldots, n$, is *i.i.d* drawn from the true distribution $\mathcal{D}$, this estimator $\widehat{Accuracy}$ in Equation (1.8) is an unbiased estimator of the true *Accuracy* as defined in Equation (1.7) [26], proved as follows:

*Proof of Unbias Estimator.*

$$\mathbb{E}[\widehat{Accuracy}] = \mathbb{E}[\frac{1}{n} \cdot \sum_{(x_i, y_i) \in T} 1_{\hat{y}_i = y_i}]$$

$$= \frac{1}{n} \cdot \sum_{(x_i, y_i) \in T} \mathbb{E}[1_{\hat{y}_i = y_i}]$$

$$\overset{i.i.d}{=} \frac{1}{n} \cdot \sum_{(x_i, y_i) \in T} \mathbb{E}_{(x,y) \sim \mathcal{D}}[1_{\hat{y} = y}]$$

$$= \frac{1}{n} \cdot n \cdot \mathbb{E}_{(x,y) \sim \mathcal{D}}[1_{\hat{y} = y}]$$

$$= \mathbb{E}_{(x,y) \sim \mathcal{D}}[1_{\hat{y} = y}]$$

$$= Accuracy$$

□

A modified double-blind visual assessment of a random sample of test sites was used to assess classification accuracy. Firstly, a random set of 304 test sites was generated across the region, and the algorithm can be found through the link: source code for random points (by Sulong Zhou). This random set corresponds to the test set $T$ with $n = 304$ and each element $(x_i, y_i)$ is *i.i.d* drawn from the true distribution $\mathcal{D}$, which is the distribution of dB value and label of the locations in the study area as shown in Figure 1.1. Next, it was assigned to a team in Nanchang who visited all accessible points from the set of 304. Then based on their experience, knowledge, and observation in both real field settings and Google Earth, they distinguished the visited sites and labeled them as water and nonwater areas. We finally verified the ground truth data by comparison with false color composite Landsat 8 imagery, and rectified 19 labels. Explicitly, this step is to assess the true label $y_i$ as water or nonwater for each $x_i$ in the test set $T$. Finally, these labeled test sites $\{(x_i, y_i, \hat{y}_i)\}_{i=1}^{n=304}$, where $\hat{y}_i = f_{\hat{\theta}}(x_i)$, were input as the ground truth information to generate a confusion matrix.

## 1.3 Results and Discussion

### 1.3.1 Confusion Matrix

The overall accuracy before test set rectification was 83.88% (Table 1.2) while the overall accuracy after test set rectification increased to 92.11% (Table 1.3). The diagonal elements (left to right, top to bottom) in the matrix represent the number of correctly classified pixels of each class, for example, the number of ground truth pixels with a label of water that was actually predicted as water during classification.

**Table 1.2:** Confusion Matrix Before Test Set Rectification.

|  |  | Predicted Label | | |
|---|---|---|---|---|
|  |  | Water | Nonwater | Total |
| Actual Label | Water | 53 | 40 | 93 |
|  | Nonwater | 9 | 202 | 211 |
|  | Total | 62 | 242 | 304 |

In contrast, the cross-diagonal elements represent misclassified pixels. A large loss of accuracy (40 out of 93) occurs at the pixels that are water in ground truth data but are classified into nonwater (Figure 1.2a). This happens for two reasons. First, many of these points are located at bound-ary pixels between two classes. Second, most points are isolated from their neighbor clusters. The boundary area has the mixed pixels problem which means both water and nonwater contribute to the observed spectral response of the pixel. In addition, the penetrating ability of C-band is unable to detect water hidden below rocks or vegetation cover where the normalized difference vegetation index (NDVI) is greater than $0.7$[31]. By contrast, there are only nine pixels that are misclassified into water while they are labeled as nonwater in ground truth data (Figure 1.2c). This is likely because the VV band can be affected by wind so that the wavy water surface will be classified into nonwater because of the diffuse refection. On the other hand, we examined each of those misclassified points and discovered that human errors in test data also compromise the overall accuracy. Twenty-two out of 40 nonwater test points and three out of nine water test points were rectified by using Landsat 8 imagery as a reference. As a result, the overall accuracy increased nearly 9%.

**Table 1.3:** Confusion Matrix After Test Set Rectification.

|  |  | Predicted Label | | |
| --- | --- | --- | --- | --- |
|  |  | Water | Nonwater | Total |
| Actual Label | Water | 75 | 18 | 93 |
|  | Nonwater | 6 | 205 | 211 |
|  | Total | 81 | 223 | 304 |

Note that the rectification of the test set $T$ does not influence the training process nor the estimation of the optimal threshold $\hat{\theta}$ provided by the Otsu Method. Since the Otsu Method is an unsupervised learning algorithm, it does not depend on the label $y$ of data for its training process. This property presents the feature of data corruption tolerance of the Otsu Method. In other words, corruption in the input data's label does not influence the actual training or performance of the Otsu Method. In addition, the test set $T$ is only used for the statistical estimation or evaluation for the performance of the classification based on the estimation of optimal threshold $\hat{\theta}$ from the Otsu

Method.

Furthermore, for factors influencing the accuracy, it is worth noting that the radar dB value with its corresponding label is linear nonseparable data [37]. In other words, there does not exist a hyperplane to clearly separate the dB value corresponding to the label of Water and the dB value corresponding to Nonwater. Because there exists two different regions or pixels $i$ and $j$, where $i \neq j$, such that they have the same dB value $x_i = x_j$, but they are actually having different label $y_i \neq y_j$, one region corresponds to water and another region corresponds to nonwater. Such linear nonseparability may decrease the accuracy of this learning algorithm, which is unavoidable because the Otsu Method is trying to use a linear threshold to separate the data. One possible example of such region $i$ and $j$ is shown in the Figure 1.2b.

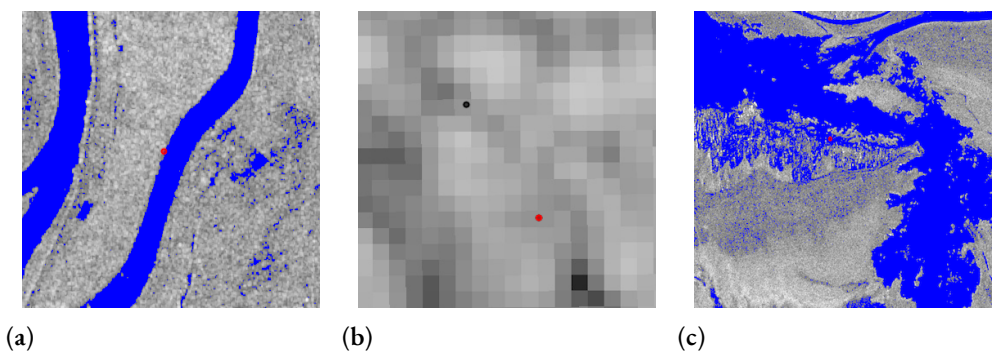

(a)                     (b)                     (c)

**Figure 1.2:** Three Situations of Misclassification: **(a)** A pixel (116.4064 E, 28.8405 N) that was misclassified into nonwater. **(b)** Two pixels with equal dB values but different truth label $y$ (116.2196 E, 28.9905 N) (116.2201 E, 28.9899 N). **(c)** A pixel (116.1584 E, 29.2136 N) that was misclassified into water.

### 1.3.2 WATER AREA

The optimal threshold for 2020 was selected as $-16.87$ through the histogram, where the low peak corresponded to water pixels while the high peak corresponded to nonwater pixels (see Figure 1.3). Similarly, the optimal thresholds for each year from 2017 to 2019 were $-14.88$, $-16.93$ and $-16.96$ respectively. Based on the auto-selected thresholds derived from the Otsu Method, the imagery was classified into water and nonwater regions. As a result, the surface water acreages of Poyang lake from 2017 to 2020 were obtained and are presented in Table 1.4 and visualized from left to right in the Figure 1.4. The surface water area decreased by nearly 650 km$^2$ between 2017 and 2018, then increased by nearly 640 km$^2$ between 2018 and 2019 and finally decreased by nearly 856 km$^2$ between 2019 and 2020. This shows that surface water area of Poyang lake decreased with fluctuation, which is consistent with other research on variation of the surface water of Poyang lake during the time period of 1988–2016[123].

In addition to the significant interannual variation, our results also showed the spatial variation of surface water area. The dry or draw down areas mainly occurred in the center and the boundary of the lake at the same time. The water areas located to the north (connected to Yangtze River) and west (connected to Gan River) accounted for most of the variation, while the water areas located at the east and south maintained much less variation.

The water area variations typically are closely associated to water level variations in Poyang lake basin[134]. Both water area and water level are dominant factors for wetlands in Poyang lake, and thus affect habitat distribution and accessibility. In this case, our classification results that show the spatiotemporal water area variations can provide robust linkages to habitat availability and suggest future research to further quantify this relationship.

**Table 1.4:** Area (km$^2$) of Poyang lake from 2017 to 2020.

| Year | Area (km$^2$) |
|------|---------------|
| 2017 | 1959.50 |
| 2018 | 1308.67 |
| 2019 | 1948.72 |
| 2020 | 1092.82 |

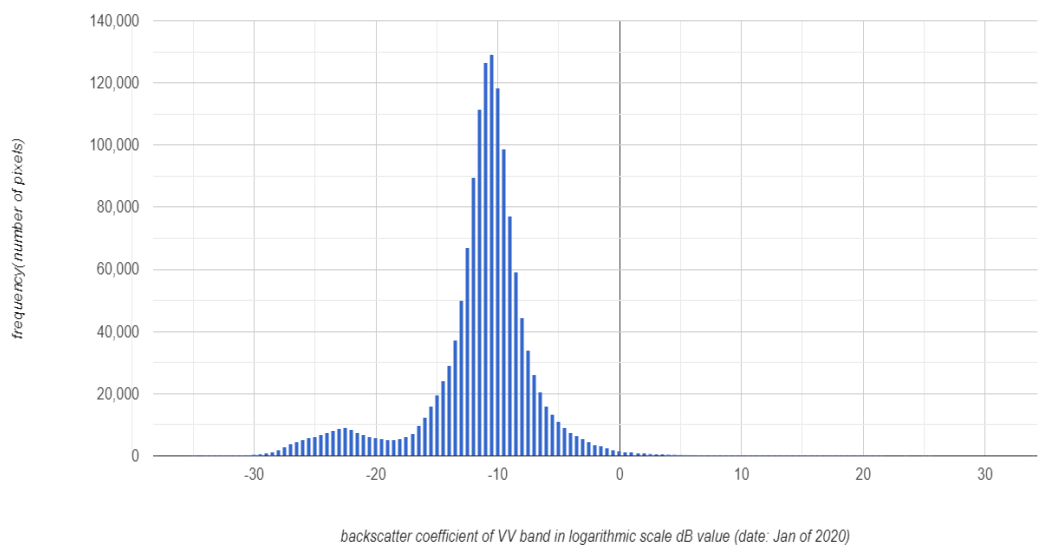

**Figure 1.3:** An Example of Histogram of VV Band in dB value (4 January 2020), the x-axis represents that backscatter coefficient is calculated in dB scale, the y-axis represents that how many pixels have the same dB value in a bin and the interval of bins is 0.5.

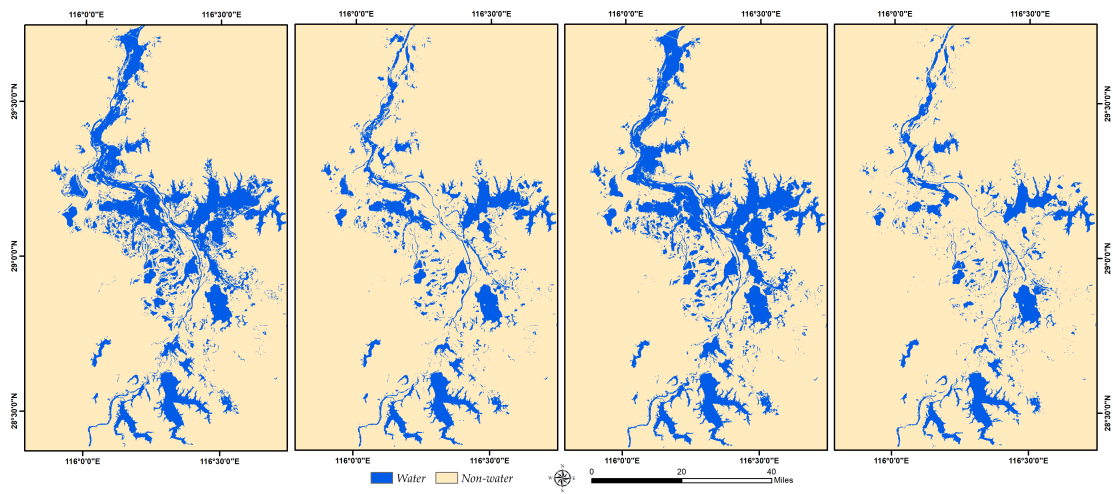

**Figure 1.4:** Inundation Area of Poyang lake in Winter of 2017, 2018, 2019 and 2020 (from left to right).

## 1.4   Conclusions

Through this research we mapped the spatio-temporal variation of Poyang lake in January from 2017 to 2020 and showed that the surface water area fluctuated annually and tended to shrink both in the center and boundary of the lake over the past four years. The variation was consistent with related Poyang lake research for earlier decades. Our mapping approach involved a novel implementation of the Otsu Method and processing of Sentinel-1 data in Google Earth Engine. GEE performed well as a powerful cloud computing platform to implement an exhaustive searching algorithm. We provided detailed mathematical explanation to enumerate the advantages and limitations of the Otsu Method that were not clearly indicated in previous remote sensing research. We also demonstrated that the Otsu Method can be an effective classifier for threshold auto-selected algorithms to extract water surface with use of Sentinel-1 data. As a result, the Otsu Method has potential to be applied to other water related studies, such as water extraction applications for other lake regions, water pollutant detection for environmental assessment and aquatic habitat mapping for ecological conservation, using the open access scripts of the threshold algorithm contributed here.

In the future, to reduce the influence of linear nonseparability nature of the data, the 2D Otsu Method[133] can be applied. However, since the Otsu Method is an unsupervised method, we have not compared its performance with supervised learning algorithms. The supervised learning algorithm requires training data that is unavailable for the January of 2020 at this time. In addition, the Otsu Method is affected by the penetration ability of single C-band radar signal so that it is difficult to capture water beneath the vegetation. To advance our research toward mapping aquatic habitat availability, we recommend the comparison between supervised and unsupervised methods by using different series of imagery to discriminate vegetation zones. This next step will allow us to identify and project the spatial distribution of available foraging habitat under varying hydrological conditions for species of concern like Siberian crane and other aquatic organisms.

**2**

# A Guided Latent Dirichlet Allocation (LDA) Approach to Investigate Real-time Latent Topics of Twitter Data during Hurricane Laura

## 2.1 Introduction

Formed on August 20 and dissipated by August 29 of year 2020, Hurricane Laura was a deadly and destructive Category 4 Atlantic hurricane[80]. The storm killed 31 people in Haiti[113], three in the Dominican Republic[85] and prompted the evacuation of more than 160,000 people in Cuba[13]. Early on August 27, Laura made landfall near peak intensity at Cameron, Louisiana[114]. It ranked with several other landfall hurricanes in a tie for 5$^{th}$ place by wind speed (150 mph) on record[69]. At least 14 people died in the U.S.[12], and it inflicted significant damage to southwestern Louisiana and southeastern Texas with insured damage estimated at close to $9 billion[34].

A key premise of making decisions for disaster response is having prompt, available and accurate situational awareness (SA)[10]. SA is "all knowledge that is accessible and can be integrated into a coherent picture, when required, to assess and cope with a situation"[99]. Retrieval of SA can help society and decision makers understand the current situation and potential hazards and forecast the ensuing risks and repercussions for the affected community[10,99]. Since relevant and timely information are necessary to inform SA, and for effective and rapid decision-making to direct response and recovery activities, the growth of sharing in-time messages through social media has contributed to provide SA[44,120].

Ranked as one of the top 10 popular social media websites, Twitter has 400 million registered users and over 500 million tweets generated every day[79]. With the capability of real-time feedback and time stamps to provide conversation updates to users, Twitter data has been used for a broad range of applications in natural disasters including fire[76], flood[87], earthquake[98], and hurricane[45]. Natural Language Processing (NLP) has been known as the most effective technology to mine tweets without fatigue and in a consistent, unbiased manner[55]. During the last decade, variant approaches for detecting the topics in a corpus of tweets have been proposed based on rapid development of neural network (NN) in NLP. Regarding to topic classification task, Convolution Neural Network (CNN) and Recurrent Neural Network (RNN) strongly outperform traditional machine learning models such as Logistic Regression (LR) and Support Vector Machine (SVM)[109,3]. However, the models based on neural network are costly and time-consuming because they require extensive training data. In addition, they are prone to underfit and overfit problems[78,33]. In particular, the pre-training step requires that millions of tweets must be labeled in advance[101,63], which impedes processing real-time tweets and thus obstructs retrieving SA information. Moreover, as high imbalance is naturally inherent in tweets of different topics[95], regular network models have not shown significant performance and extra complicated strategies are in demand to address the imbalanced data problem[61].

Latent Dirichlet Allocation (LDA) model was introduced to reduce the workload and handle imbalanced data[77,112] with a bonus to investigate latent topics of a corpus[5]. In general, LDA models assume that documents consist of a distribution of topics and that topics are made up of a semantically coherent distribution of words. It is an unsupervised algorithm that models each document as a mixture of topics in order to generate automatic summaries of topics in terms of a discrete probability distribution over words for each topic, and further infers per-document discrete distributions over topics [15]. As such, this work intends to leverage LDA models to rapidly assist in disaster response and inform SA, and provide reusable code for future implementation.

Specifically, the contributions of the present work are:

1. Propose a guided LDA approach that integrates domain knowledge, coherence models, latent topics visualization, and validation from official reports.

2. Mine tweets to reveal common classification schema for future use in supervised model.

3. Investigate temporal latent topics to further inform SA for decision makers and local citizens during the Hurricane Laura.

## 2.2   RELATED WORK

Previous research has proposed different topics classification schema for different disaster events. Vieweg *et al.*[119] proposed 12 general topics schema for fire and flooding events including warning, preparatory activity, hazard location, flood level, weather, wind visibility, road conditions, advice, evacuation information, volunteer information, animal management, and damage/injury reports. Imran *et al.*[54] proposed five general topics schema for Tornado Joplin and Hurricane Sandy including caution and advice, information source, donation, casualties and damages, and unknown. Huang and Xiao[51] proposed a 47 topics schema based on four different phrase stages: mitigation, preparedness, emergency response, and recovery. In fact, existing studies indicated that the design of the coding schema and associated message classification methods are highly varying and depend on different disaster event types , the analysis purpose or classification purpose (CP) of the event, and the social media platforms[120]. Since the classification schema varies across research studies and disaster events, it is necessary to identify our own topics schema for Hurricane Laura.

Supervised learning has been used for extracting information from social media for disaster management and response. Habdank et al.[46] collected 3,785 pieces tweet data during an pipeline explosion accident in Ludwigshafen, Germany, and applied supervised learning algorithms, including Naive Bayes, Decision Trees, Random Forests, Support Vector Machines and Neural Networks, to classify whether a piece of tweet was relevant to this accident. Pouyanfar et al.[93] analyzed the audio and visual content from more than 1,000 video clips from Hurricane Harvey and over 450 video clips from Hurricane Irma on YouTube, with using Neural Networks to classify these videos into seven different semantic classes. In these studies, large sets of pre-labeled data were required for the training and validation process, and the label task was extremely time consuming. Also, each piece of data was only assigned to a single label. However, a social media message can often contain multiple semantic meanings. For example, a message "winds are still 65mph...over 10 people killed and avoid the flooding area" contains information in "Information Source", "Advisory" and "Casualty" categories. Furthermore, the categories and the criterion of each category defined in these studies are event dependent, which potentially limits their applications across different events.

LDA has been applied to several fields such as software engineering, political science, medical science, geography, etc. More than 200 related scholarly articles from 2003 to 2016 have been fully discussed and reviewed to discover the development, trends and intellectual structure of topic modeling based on LDA[60]. The prominence of application of LDA model application continues on research regarding natural disasters. Sit et al.[106] proposed a NLP model with LDA to to extract detailed information such as affected individuals, donations and support, caution and advice from tweet content for Hurricane Irma. Karami et al.[64] proposed a Twitter Situational Awareness framework with LDA model to track the negative concerns of people during the 2015 South Carolina flood. Alam et al.[2] proposed an unified framework including LDA to help humanitarian organisations in their relief efforts during Hurricanes Harvey, Irma, and Maria. However, there remain two major problems. First, these studies lack a quantitative validation to evaluate the coherence of topic clusters derived from the LDA model. Second, regular LDA models are limited to short tweet text because frequent words are shared between topics and sometimes distinct topics cannot be tagged for each group of words[122].

Evaluation of LDA models is notoriously challenging due to its unsupervised training process. Based on different metrics and purposes, the common evaluation methods include human judgment (eye balling), intrinsic evaluation (perplexity and coherence), and extrinsic evaluation[96]. In the original LDA paper, Blei *et al.*[15] recommended the perplexity metric as an intrinsic evaluation to justify the selected LDA model. However, subsequent studies have revealed that the perplexity is not correlated to, and even sometimes slightly negatively correlated to human judgment. Sievert and Shirley[104] developed LDAvis, a web-based interactive visualization python package, to flexibly conclude a fitted LDA model. It allowed for a deep inspection of the keywords most highly associated with each individual topic but without showing specific quantified values as validation. Roder *et al.*[97] proposed the concept of six topic coherence metrics by measuring the degree of semantic similarity between high scoring words in the topic. However, the score only reflects the quantified results but does not demonstrate human interpretable content for each topic.

To solve above problems, we proposed a guided LDA approach with pre-defined topic candidates based on domain knowledge, the most effective coherence metric and topic visualization tool. This approach allows us to select optimal topic numbers and validate the results with both intrinsic evaluation and human judgments. Furthermore, to our knowledge, no other research has used the LDA model with our guided approach to inform SA for the latest Hurricane Laura.

## 2.3 Methods

Our framework is mainly comprised of four steps: data collection, data preprocessing, guided LDA and latent topic clusters selection.

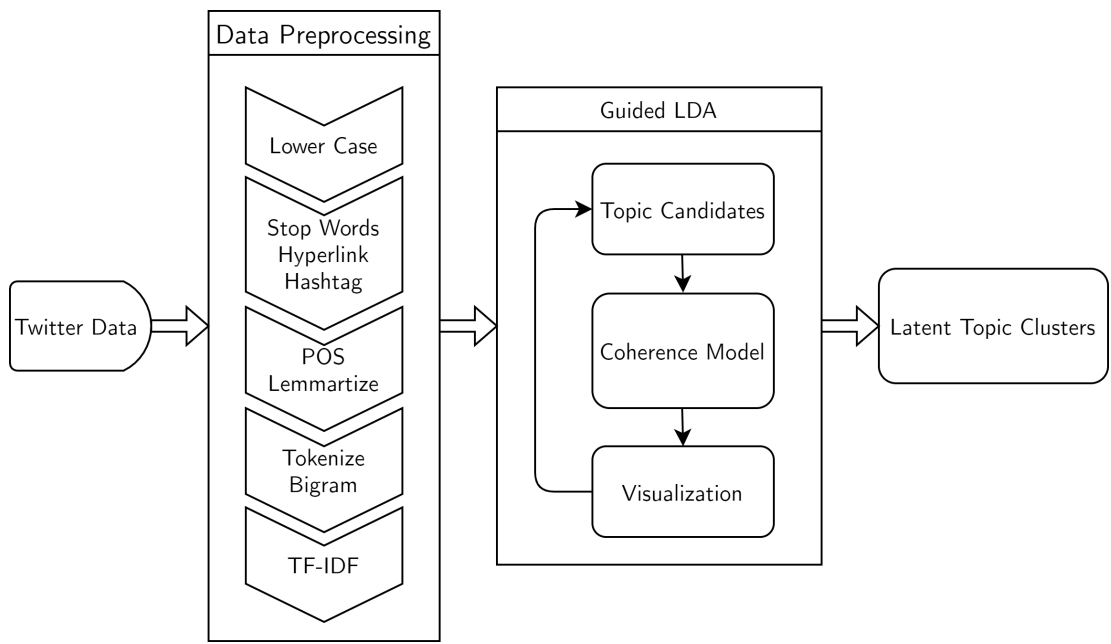

**Figure 2.1:** Overview of the proposed workflow of Guided LDA model

### 2.3.1 DATA COLLECTION

Twitter API, despite being free, is limited as users can only retrieve a small portion of relevant tweets. Twint[92] is an advanced Twitter scraping tool written in Python that allows for scraping Tweets from Twitter profiles without using Twitter's API. Twint utilizes Twitter's search operators to let you scrape Tweets from specific users, scrape Tweets relating to certain topics, hashtags and trends, or sort out sensitive information from Tweets.

### 2.3.2 DATA PREPROCESSING

Preprocessing is a requisite to convert tweets into a form that is predictable and analyzable for our task. Here are some of the approaches we used in this project.

- Lowercasing: removes duplicate words and significantly helps with consistency of expected

output, e.g. Hurricane → hurricane.

- Part of Speech (POS) Tagging: assigns a POS tag, such as noun, verb, adjective, etc, to each token depending on its usage in the sentence. It is essential for building lemmatizers, e.g. storm → noun, kill → verb.

- Lemmatization: removes inflections and maps a word to its root form without changing its POS, e.g. worse → bad, damaging → damage.

- Stopword Removal: Stop words are a set of commonly used words in a language and across tweets, e.g. "a", "the", "is", "are", "hurricane" and etc. By removing low information words from text, we can focus on the important words instead.

- Noise Removal: removes characters, digits, hashtags, hyperlinks and pieces of text that can interfere with our text analysis.

- Tokenization: separates a piece of tweets into smaller units such as words, or a pair of words (2-gram), e.g. "Mass destruction in one Lake Charles neighborhood" → "Mass", "destruction", "in", "one", "Lake_Charles", "neighborhood".

- Term Frequency — Inverse Document Frequency (TF-IDF): assigns a weight to each word based on words frequency which balances the importance of the word in the tweets and corpus.

### 2.3.3 Topic Model Analysis

#### Regular LDA

LDA is a generative probabilistic model. We assume the total amount of topics as $k$. The probability of a generated collection $D$ of tweets $d$ is expressed as[15]:

$$p(D|\alpha,\beta) = \prod_{d=1}^{M} \int p(\theta_d|\alpha) \cdot \left(\prod_{n=1}^{N_d} \sum_{z_{dn}} p(z_{dn}|\theta_d) p(w_{dn}|z_{dn},\beta)\right) d\theta_d \qquad (2.1)$$

Where:

- $D$ is the collection of tweets $d$ and composes of $M$ tweets, $d = 1, \ldots, M$.

- $\theta_d$ is the distribution of the $k$ topics for a tweet $d$, where $\theta_d$ is a $k$ dimensional vector and $\sum_{i=1,\ldots,k} \theta_d^i = 1$. Specifically, $\theta_d^i$ means the probability for tweet $d$ to have the topic $i$. Furthermore, $\theta_d$ is selected through a topic-tweet distribution $p(\theta_d|\alpha)$, which has a prior Dirichlet Distribution with parameter $\alpha$, i.e. $\theta_d \sim p(\theta_d|\alpha) = Dir(\alpha)$.

- For the tweet $d$, it has $N_d$ amount of words. For the $n$-th word $w_{dn}$ in the tweet $d$, where $n = 1, \ldots, N_d$, it has possible topic $z_{dn}$.

- The probability or word-topic distribution for the word $w_{dn}$ given the topic $z_{dn}$ is described with parameter $\beta$, i.e. $p(w_{dn} = j|z_{dn} = i, \beta) = \beta_{ij}$.

- The conditional probability for topic $z_{dn}$ to be $i$, given the topic distribution $\theta_d$, is $p(z_{dn} = i|\theta_d) = \theta_d^i$.

The prior distribution for the topic-tweet distribution and the word-topic distribution are given, i.e. $\alpha$ and $\beta$ are initialized as hyper-parameters. During the training process, the goal is to train the posterior distribution of these two distributions based on the observed collection of tweets $D$, such that we can maximize the generative probability for $p(D|\alpha,\beta)$.

At the end of training, the semantic meaning of each topic $i$ is represented by the word-topic distribution associated with topic $i$, i.e. the $i$-th row $\beta_i$ of the matrix $\beta$. The conditional probability $\beta_{ij}$ for word $j$ based on the topic $i$ is the semantic weight for this word $j$ under this topic $i$. The higher weight of a word means that this word is more representative for the topic. Therefore, we can select the top 10 words with the highest weights in the row $\beta_i$ to describe the semantic meaning for this topic $i$.

## Coherence Model

The coherence value for a single topic measures how the top scoring words in this topic are semantically similar to each other. As compared to the conventional topic perplexity measurement, which measures how well a trained model can fit or represent the distribution for an unseen tweet[65], coherence measurement focuses more on the interpretative aspect of a topic and has a high correlation to the human scoring of topics[97], whereas the perplexity measurement is sometimes not related to and may even be negatively to human judgement and interpretation[57]. Therefore, coherence value can provide a better measurement on the semantic performance of the LDA model and the selected topics. Furthermore, due to the correlation between the coherence measurement and human judgement, the higher coherence value indicates that the topic is more interpretive and meaningful for human interpretation. Thus, we adopt the coherence measure during the process of hyper-parameter tuning and select the hyper-parameter $k$, $\alpha$ and $\beta$ at which the highest coherence value is reached. In this way, the selected hyper-parameters $k$, $\alpha$ and $\beta$ can provide a more meaningful topic selection for tweets along with the LDA model, compared to a random selected hyper-parameters.

## Topic Visualization

pyLDAvis is designed to help users interpret the topics in a topic model that has been fit to a corpus of text data [104]. The package extracts information from a fitted LDA topic model to inform an interactive web-based visualization.

Since each topic is embedded in a high dimensional space, the pyLDAvis applies the mutidimensional scaling and dimensionality reduction techniques to project each topic's high dimensional embedding onto the 2$D$ space for visualization purposes [104]. Specifically, we use the t-distributed stochastic neighbourhood embedding (t-SNE) method in the pyLDAvis package, which is a nonlinear dimensionality reduction method. As compared with pyLDAvis' default Principle Component Analysis (PCA) method, which maximizes the variance of each topic's projection along the new axis [124], the t-SNE method considers the relative similarity between each topic in the high dimensional embedding and preserves this similarity after the projection [71]. Therefore, the t-SNE method outperforms the PCA method for visualizing the relative relationships between each topic in the projected 2$D$ space and provides a better measurement for the quality of the selected topics.

Compared to traditional clustering techniques where each tweet can only belong to single topic, an advantage of pyLDAvis is that a word can be clustered to different topics [104,2]. For example, the word 'hit' can appear in a context regarding to information source or damage, and the word 'evacuate' can appear in a context regarding to advisory or relief. In this case, it can better present the nature of language.

## LDA with Guided Approach

The Guided approach initially defines topic candidates based on previous research and domain knowledge. The topic candidates provide a specified direction for the word-topic distribution and the topic-tweet distribution to converge toward during the training of the LDA model.

In the regular LDA model, the model is purely trained with a probability based target function in equation 2.1. It captures the relationships between words based on the frequency. Therefore, this model may only capture the superficial relationships between the words, which are the most frequent and apparent in the collected tweets[58]. However, this probability-based model performs badly for those less frequently appearing patterns of tweets, because lower frequency means that these patterns contribute less to the target function and thus become less important to be considered by the model. This model will merge these less frequent topics together into a larger topic group instead. Even though this merging process benefits a higher value for the probability-based target function, it harms the coherence and semantic of the topics, as two semantically disparate topics are merged together. Especially in the tweet data, the distribution of different topics is imbalanced and the regular LDA model will generate less interpretative topics.

Thus, we used a guided approach to solve this problem with human intervention. The potential range of numbers and contents of topic were estimated by integrating quantitative (coherence value) and qualitative evaluation (topic visualization). Then the event-specific SA categories and corresponding dominated words were manually summarized with minimal efforts.

## 2.4 Results and Discussion

### 2.4.1 Effective Tweets Count and Rate

We collected the tweets with removing duplicate retweets and searching criterion: 'Hurricane Laura OR Hurricane OR Laura' across 10 days between Aug 21 2020 and Aug 30 2020. This time frame was selected based on the landing and duration of Hurricane Laura from Aug 25 2020 to Aug 29 2020. After data pre-processing, some irrelevant tweets were filtered out (Table 2.1) and the number of tweets highly related to Hurricane Laura per day ranged from $1,184$ tweets to $59,921$ tweets (Table 2.2). On average, $50.85\%$ tweets per day were filtered out for the input for our LDA model.

**Table 2.1:** Examples of irrelevant tweets.

175,000 dead from the virus, over 1 million unemployed, no money, losing homes. California is on fire. Louisiana-Texas are preparing to be hit with double hurricanes and here you are once again pretending nothings going on and you, going to have a good old time at a party. Again WTF (8.23)

Texas is facing a double hurricane, while California faces wildfires yet again. What is the republican plan to combat climate change? That what I like to hear. (8.25)

As if the US does not have enough to deal with now we have the consequences of climate change, Horrific wildfires and a Cat 5 unsurvivorable Hurricane and surge heights, during a pandemic. We must come together despite ethnicities, skin color and politics to survive. (8.26)

**Table 2.2:** Summary of tweets before and after preprocessing per day during Hurricane Laura

| Date | # of Tweets before preprocessing | # of Tweets after preprocessing | effective percentage |
|------|----------------------------------|----------------------------------|----------------------|
| 8.21 | 2972 | 1184 | 39.84% |
| 8.22 | 28359 | 12779 | 45.06% |
| 8.23 | 35911 | 16867 | 46.97% |
| 8.24 | 25842 | 12363 | 47.84% |
| 8.25 | 48794 | 32697 | 67.01% |
| 8.26 | 128292 | 59921 | 51.57% |
| 8.27 | 61339 | 27466 | 44.78% |
| 8.28 | 15582 | 6910 | 44.35% |
| 8.29 | 9821 | 5114 | 52.07% |
| 8.30 | 6631 | 3026 | 45.63% |
| *Total* | 363542 | 184564 | 50.76% |

As shown in the Figure 2.2, the temporal pattern of tweets frequency closely corresponded to the temporal pattern of Hurricane Laura. As soon as Hurricane Laura intensified into a tropical storm on Aug 21, 1, 184 highly related tweets emerged. With a rapid intensification of Hurricane Laura on August 26 [114], the amount of tweets rapidly increased to 59, 921 in the same day (around 2.8 times than amount of tweets in previous day). During the 10-day period of Hurricane Laura, Twitter contributed 184, 850 related unique tweets in total. However, the average effective rate of tweets was around 51%, which meant only half of the tweets truly contributed to related topics on SA after data pre-processing. The lowest percentage was 39% (1, 184 out of 2, 972) on Aug 21 while the highest percentage was 67% (32, 697 out of 48, 794) on Aug 25. Since tweets contained a significant amount of noise comprised of fake news, robot messages, advertisements and irrelevant cross-event topics [72,128], pre-processing data prior to topics analysis was necessary and crucial to retrieving precise situational information.

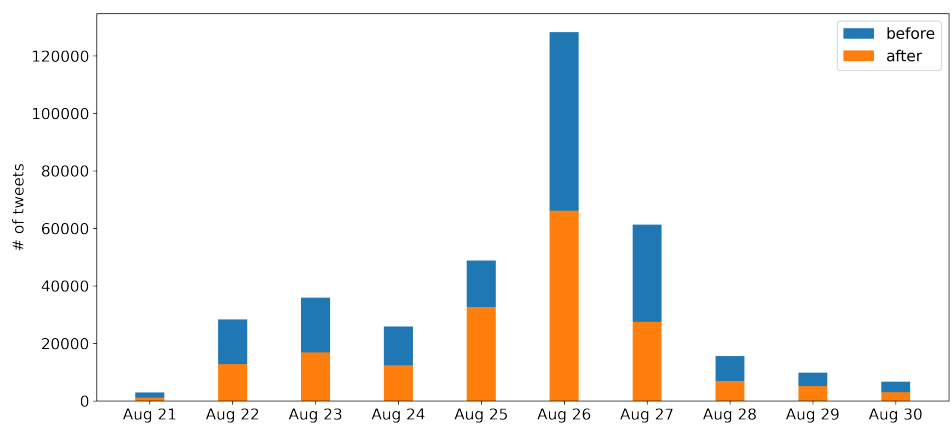

**Figure 2.2:** Histogram of tweets count before and after preprocessing per day during Hurricane Laura

Similarly in previous research, 217, 074 negative tweets, or 25% of original data, were filtered to showcase helping effective disaster response during the 2015 South Carolina flood because of the Hurricane Joaquin[64], and 20% of over millions of tweets with downloadable images were used as complementary information to improve situational awareness with case studies of Hurricanes Harvey, Irma and Maria[2]. The large amount of data, yet with low percent effectiveness, can illustrate the temporal evolution of the hurricane events, thereby informing situational awareness and assisting in disaster response.

### 2.4.2   Optimal Daily K Topics

We designed a 4-step workflow to estimate the optimal K latent topics for each day. In the first step, we summarized the five most common situational awareness topics including 'Advisory', 'Casualty', 'Damage', 'Relief' and 'Information source' (Table 2.3) from previous hurricane related research[119,54,51,64,2]. In our second step, we set different hyper-parameters in a regular LDA model, and then evaluated each model based on coherence value. The hyper-parameters included topic numbers k, $\alpha$ and $\beta$ in the LDA Equation 2.1. For each combination of them, we trained with 100 iterations to generate the coherence values and repeated this training process with 10 trials to calculate the average of coherence values (Figure 2.3). Based on the coherence value across the 10 trials, we selected the pair of $(k, \alpha, \beta)$ corresponding to the first peak on the graph (Table 2.4). However, for certain days, there might not exist a clear peak on the graph, as shown in Figure 2.3. In this case, instead of choosing a specific value for the number of topics $k$, we chose a range from 5 to 8 for $k$. In the third step, using the selected hyper-parameters from the second step, we visualized the generated topics with pyLDAvis (Figure 2.4), evaluated if the current topics were humanly interpretable, and added two additional category candidates: 'Emotion' and 'Animal'. Lastly, we manually selected key words from the top 30 words visualized from pyLDAvis and matched them with the candidates category in Table 2.3. The final optimal topic number is shown in the Figure 2.5, and corresponding key words for each day are shown in the Table 2.5.

In the second step, the coherence value (Figure 2.3) and the optimal amount of topics were in-fluenced by the value of $\alpha$ and $\beta$, where $\alpha$ was related to the distribution of topics given tweets and $\beta$ was related to the distribution of words given topics. Empirically, a larger value of $\alpha$ represents more topics that a tweet can contain and it is usually initialized with the value of $\frac{1}{k}$, where $k$ is the total amount of potential topics. In our model, we observed the optimal value of $\alpha$ was 0.1. On the other hand, a larger value of $\beta$ showed that a topic consisted of more words. Based on the observation of coherence value across different pairs of hyper-parameters, we mainly selected the value of 0.6 for $\beta$ on most of days, except for the value of 0.9 on 8.26 and 0.3 on Aug 27 (Table 2.4).

**Table 2.3:** Topic candidates referenced from previous research and supplemented from our model.

| Candidates | Stage | Description |
|---|---|---|
| Advisory | pre and during | cautions, advices, warnings, alerts and preparedness |
| Casualty | during and post | missing, injured, and/or dead people |
| Damage | during and post | impacts, damages and affected industries and areas |
| Relief | during and post | services, donations and fundraisers to disaster response |
| Information source | pre, during, and post | messages from an official news source, media |
| Emotion | pre, during, and post | public concerns and feelings |
| Animal | during | pet and wildlife protection |

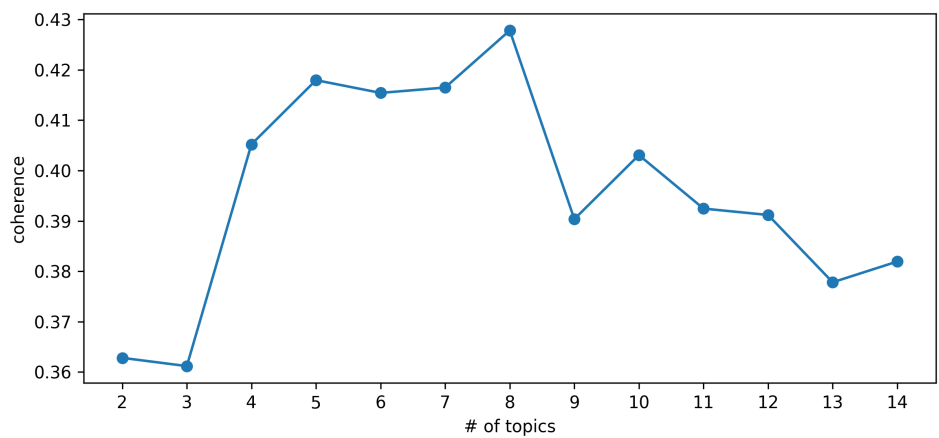

**Figure 2.3:** Coherence values of different K selected topics on Aug 26

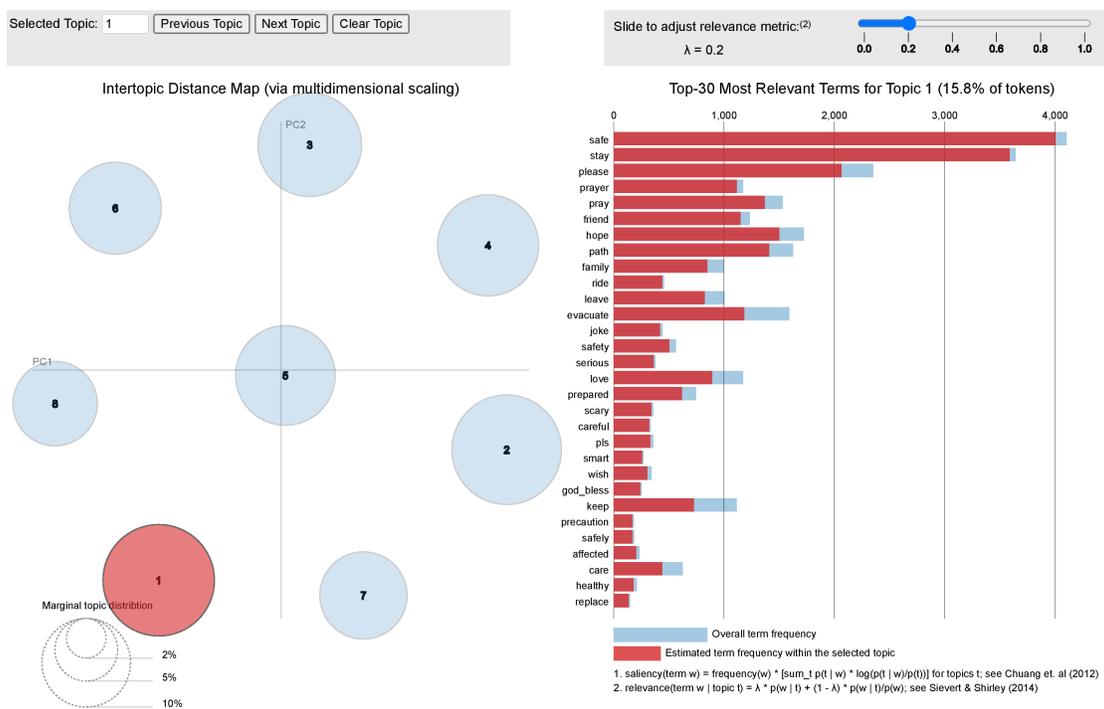

**Figure 2.4:** An example to visualize topic clusters and Top-30 dominant words for Topic 1 on Aug 26

In the third step, not all the visualizations of k topics for each day were presented because of space limitations. Instead, we used the visualization for Aug 26 as an example (Figure 2.4). On the left, the Inter-topic distance map depicts the relative position between each topic under this 2D projection space, which was also the relative position of these topics in their original high dimensional embedding space, under the t-SNE method. Specifically, the closer positions of two topics under this 2D space meant the more similarity of these two topics and the farther positions represented the more exclusiveness of these two topics. On the right, the bars demonstrated the top 30 most relevant words (i.e. words with a high probability of being associated to a topic) for the most prevalent topic among eight topics. The red bars showed word/term frequency for a specific topic while the blue bars indicated the term frequency for all the topics. Figure 2.4 depicts that 'safe','stay','please', 'hope' and 'pray' were dominant words for this topic (long red bars) while they were not dominant words for other topics (short blue bars). $\lambda$ determined the weight given to the probability of a word under a topic relative to its lift * (measuring both on the log scale). Setting $\lambda = 1$ resulted in the familiar ranking of words in decreasing order of their topic-specific probability, and setting $\lambda = 0$ ranked terms solely by their lift [104].

**Table 2.4:** Tuned hyperparameters($\alpha, \beta, k$) per day during Hurricane Laura

| Date | $\alpha$ | $\beta$ | range of topics with Coherence Value | # of topics with pyLDAvis |
|------|------|------|--------------------------------------|---------------------------|
| 8.21 | 0.1 | 0.6 | $[3, 4]$ | 3 |
| 8.22 | 0.1 | 0.6 | $[4, 5, 6]$ | 4 |
| 8.23 | 0.1 | 0.6 | $[5, 6, 7, 8]$ | 6 |
| 8.24 | 0.1 | 0.6 | $[6, 7, 8]$ | 6 |
| 8.25 | 0.1 | 0.6 | $[6, 7, 8]$ | 6 |
| 8.26 | 0.1 | 0.9 | $[5, 6, 7, 8]$ | 8 |
| 8.27 | 0.1 | 0.3 | $[6, 7, 8]$ | 6 |
| 8.28 | 0.1 | 0.6 | $[3, 4, 5, 6, 7]$ | 6 |
| 8.29 | 0.1 | 0.6 | $[5, 6, 7, 8]$ | 5 |
| 8.30 | 0.1 | 0.6 | $[2, 3, 4]$ | 3 |

As shown in Figure 2.5, the temporal pattern of topic numbers per day increased for the first seven days while it dropped for the last three days. The greatest number of topics were discussed on Aug 26 and Aug 27. Specifically, the topic of 'Advisory', 'Relief', 'Information source' and 'Emo-tion' were mainly discussed in the pre-landfall stage of Hurricane Laura. Starting from Aug 23, the topic of 'Casualty' and 'Damage' appeared in the tweet. On Aug 26 and Aug 27, the topic of 'Animal' was also discussed along with all other topics. In the post stage of Hurricane Laura, the discussion about 'Advisory' decreased and focused more on 'Casualty', 'Damage', 'Relief', 'Information source' and 'Emotion' (in the Table 2.5).

---

*$lift = P(topic \wedge word)/(P(topic) \cdot P(word)) = P(word|topic)/P(word)$

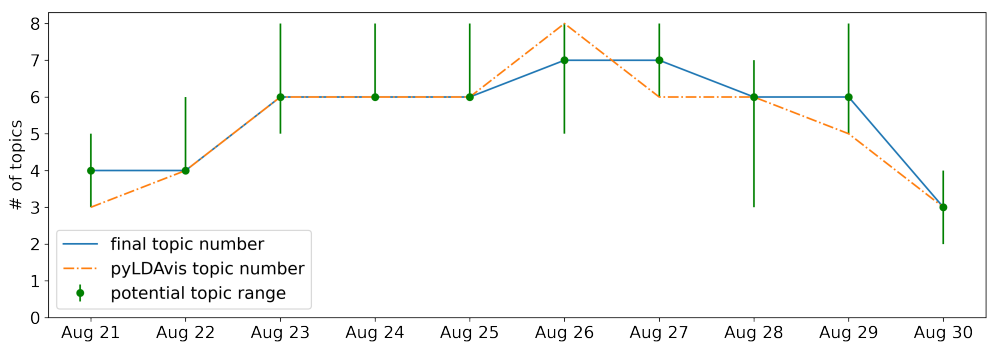

**Figure 2.5:** Potential topic range, pyLDAvis topic numbers and final topic numbers per day

### 2.4.3 TEMPORAL TOPIC CONTENT

To verify how efficiently and precisely our models reflected the real time latent topics during Hurricane Laura, we designed a 2-step verification. First, we retrieved $2 - 5$ representative tweets for each topic. Second, we compared the tweets with official reports published by government, authoritative organization and news media. As shown in Table 2.7 - 2.11, the contents of tweets regarding advisory, casualty, damage, relief, and information source were verified by public announcements from Department of Energy (https://www.energy.gov/), Natural Hurricane Center (https://www.nhc.noaa.gov/), Federal Emergency Management Agency (https://www.fema.gov/), Louisiana Department of Health (https://ldh.la.gov/), American Red Cross (https://www.redcross.org/), and etc.

**Table 2.5:** Final topic categories and corresponding keywords per day during Hurricane Laura (8.21 - 8.25)

| Date | Advisory | Casualty | Damage | Relief | Information source | Emotion | Animal |
|------|----------|----------|--------|--------|--------------------|---------|--------|
| 8.21 | warn, stock, stay safe, preparedness | | | shelter, help, protect, evacuate | eye, rain, flood, fema, youtube | scream, worried | |
| 8.22 | plan, warning, stay safe, prepared, public advisory, local statement | | | help, give, supply | cone, shear, tornado, flood, update, intensity, category, twitter, | worry, surprised | |
| 8.23 | prep, advisory, stay safe, warning issue, | die | power, collide, knock utility, terrebonne, lafourche, bernard | service, supply, evacuate, ascension | mph, surge, tornado, heavy rain, weather channel, weather authority, twitter | god, hope, care, pray, bless, scary | |
| 8.24 | tip, prep, stock, warning, checklist, stay safe | kill, die, deadly | ruin, wipe | evacuation, responder, supply, amazon | wind, heavy rainfall, cancel class, nhc, twitter, free fl511 | god, pray, hope, scary | |
| 8.25 | stay, safe, prepare, plan, ready | kill, death, | loss, devastate, break, affected | food, money, evacuate | flood, tornado, school close, nhc, news, popular google | hope, pray, care | |

**Table 2.6:** Final topic categories and corresponding keywords per day during Hurricane Laura (8.26 - 8.30)

| Date | Advisory | Casualty | Damage | Relief | Information source | Emotion | Animal |
|------|----------|----------|--------|--------|--------------------|---------|--------|
| 8.26 | stay, safe, careful, prepare, precaution, public advisory | die, kill, dead, death, | lng, oil price, power, trap, slam, house, catastrophic, sink entire, lake charles, beaumont | map, money, resource, recovery, evacuate, evacuation order, maximum sustain | gust, typhoon, 115 mph, forecast, emergency, flash flood, rapidly intensify, fox news, weather channel | cry, amen, hate, hope, pray, stress, terrify, love, peace, god bless | mare |
| 8.27 | warn, calm, careful, stay safe, | death, kill, unsurvivable. | slam, beast, outage, devastate, catastrophic, sulphur | map, radar, rescue, shelter, donate, protect, community | eyewall, gust, nhc, noaa, twitter, youtube, fox news abc news | hope, amen, jesus, mercy, peace, prayer, god bless | pet, mare, animal |
| 8.28 | safety, advisory, prepare, poison, carbon monoxide, | victim, survivor, death toll | oil, batter, topple, refining, pollutant, gas industry, chemical plant, mississippi river, flow backward, destructive path | fund, donate, rebuild, support, response, recovery, voluntary, 90999 food, aerial picture | 150 mph, emergency, flash flood, npr, nyt, wnt, fema, abc news washington post | hope, pray, relief, | |
| 8.29 | safety, poison, prepare, generator, carbon monoxide | death toll, woman dead, | topples | fund, service, donation, recovery, cleanup start, | rain, tornado, remnant, nyt, twitter, youtube | care, pray | |
| 8.30 | | victim, death toll, family perish | power, loss value, devastate, destruction | donate, pledge, support, sell ebay, recovery | | | |

**Table 2.7:** Tweets example of Topic 'Advisory' and official reports

| Tweets | 1. Hope everyone whose getting hit by the hurricanes stay safe!!! Stock up on the important things and charge your phones. (8.21) |
|---|---|
| | 2. Important PSA but PLEASE DO NOT RUN A GENERATOR IN-SIDE EVER!! OR IN ANY ENCLOSED SPACES!!! YOU WILL DIE FROM CARBON MONOXIDE POISONING. About half of the confirmed deaths so far from hurricane laura were bc of improper generator usage!! PLEASE DO NOT RUN GENERATORS INDOORS. (8.28) |
| Report | Department of Energy posted a public advisory regarding to "Using Portable/Emergency Generators Safely" and warned that using the generator incorrectly can cause carbon monoxide poisoning[35]. |

**Table 2.8:** Tweets example of Topic 'Casualty' and official reports

| Tweets | 1. BBC News - Hurricane Laura death toll climbs to 14 in the US. (8.28) |
|---|---|
| | 2. Hurricane Laura victims may go weeks without power, deaths climb to 14 via @ABCNews (8.28) |
| Report | Louisiana Department of Health verified 14 deaths tied to Hurricane Laura on Aug 30[67]. |

**Table 2.9:** Tweets example of Topic 'Damage' and official reports

| Tweets | 1. Keeping an eye on the city camera from our sister station KPLC in Lake Charles, you can see lights going out, outages spreading as Hurricane #Laura pushes into southwest Louisiana. Nearly 20,000 without power now, and that number will only grow tonight. (8.26) <br> 2. Updated at 11:45 a.m. Hurricane Laura tore through a region that is home to dozens of major oil refineries, petrochemical plants and plastics facilities. #atx #austin #all512 (8.28) |
|---|---|
| Report | 1. As of 8:30 AM EDT, September 25, there were approximately 15,000 customer outages reported across Louisiana and Alabama[28]. <br> 2. Hurricane Laura caused significant damage to transmission infrastructure in portions of Louisiana and Texas[28]. <br> 3. National Public Radio (NPR) reported that millions Of pounds Of extra pollution Were released because Hurricane Laura shut down dozens of oil refineries, petrochemical plants and plastics facilities[84]. |

**Table 2.10:** Tweets example of Topic 'Relief' and official reports

| Tweets | 1. Today, Red Cross Health Services volunteer, Karen Watt, set off for Baton Rouge, LA to support families impacted by hurricanes Marco & Laura. Karen is one of a dozen volunteers from North Texas who has left the region to assist. Learn how you can help (8.25) <br> 2. We committing a $1M donation to relief efforts across Louisiana & Texas for those affected by #HurricaneLaura. If YOU want to help, text LAURA to 90999 for @RedCross, or FOOD to 80100 for @WCKitchen, and $10 will be added to your Verizon Wireless... (8.28) <br> 3. Residents in southwestern Louisiana embarked Saturday on the epic task of cleaning up after Hurricane Laura tore through parts of the state. (8.29) |
|---|---|
| Report | 1. The American Red Cross announced the relief information including food, water and other volunteer assistance, and Red Cross Annual Disaster Giving Program (ADGP) members, such as Amazon and American Airlines[9]. <br> 2. Verizon Foundation commits $1 million to Hurricane Laura relief efforts[59]. |

**Table 2.11:** Tweets example of Topic 'Information source' and official reports

| | |
|---|---|
| Tweets | 1. FL511 reminds motorists to create a hurricane preparedness plan that includes recommendations in the event of an evacuation during COVID-19. Download the free FL511 Mobile App to know before you go. #FL511 (8.24) |
| | 2. Hurricane Laura reaches extremely dangerous Cat 4 strength, NHC warns of unsurvivable storm surge (8.26) |
| | 3. An extreme wind warning is continues for Beaumont TX, Lake Charles LA, Port Arthur TX until 1:00 AM CDT for extremely dangerous hurricane winds. Treat these imminent extreme winds as if a tornado was approaching and move immediately to an interior room or shelter NOW!. (8.26) |
| | 4. Hurricane Laura is appraoching the southwest LA with 150 mph winds and an expected storm surge of 20 ft. Laura will move quickly north after landfall and decrease in strength significantly. (8.26) |
| | 5. New imagery from @NOAA reveals devastation wrought by Hurricane #Laura 1/ Grand Chenier, Louisiana, near where the highest surge was recorded (8.28) |
| Report | 1. National Hurricane Center posted daily update on wind speed of Hurricane Laura since Aug 21. On Aug 25, it announced that maximum sustained winds have increased to near 90 mph (150 km/h) with higher gusts[83]. |
| | 2. From August 27 to 31, the National Geodetic Survey (NGS) collected aerial damage assessment images in the aftermath of Hurricane Laura. Imagery was collected in specific areas identified by NOAA in coordination with FEMA and other state and federal partners[82]. |

**Table 2.12:** Tweets example of Topic 'Emotion' and official reports

| Tweets | 1. Taking a break from the RNC to get in my car to go to the grocery store for the first time in 7 months because I been too scared of being killed by a mundane daily task, but now two hurricanes are coming and I more scared of running out of water and food than the virus (8.24) 2. my first hurricane in Houston & I, scared after the email I got from the apartment office I just hope water doesn't get into my car (8.25) 3. Pray for the people of Southwestern Louisiana and eastern Texas Gulf coast who are desvated by the aftermath of Hurricane Laura and for those who died. (8.27) |
|---|---|
| Report | Intercessors for America started a survey to ask for payer for Hurricane Laura. The comments indicated most of the people were worried about the "unsurvivable" storm and prayed day and night[53]. |

**Table 2.13:** Tweets example of Topic 'Animal' and official reports

| Tweets | 1. FREE LARGE ANIMAL STALLS AVAILABLE at Mississippi Fairgrounds as temporary shelter for evacuated horses and livestock affected by #HurricaneLaura. Call 601-961-4000 for more info. @MSDeptofAg @MSEMA @CommAndyGipson @GregMichelMSEMA @WJTV @halterproject #DisasterAnimals (8.27) 2. The North Shore Animal League in Port Washington is accepting donations for food and supplies to help pets displaced by Hurricane Laura. (8.27) |
|---|---|
| Report | 1. On Sep 9, it was reported that the state fair was canceled this year, but livestock barns and horse stalls at the Oregon State Fairgrounds got some use after all. Volunteers remained at the fairgrounds helping care for displaced animals. There were photos of rabbits, horses and goats attached in the news[70]. 2. North Shore Animal League wrote a blog about "Supplies Needed for Partner Shelter in Path of Hurricane Laura" to ask for donations by 5pm Friday, August 28[6]. |

The authoritative evidence not only showed the credibility but also indicated the timeliness of the tweets that we filtered out. Among these temporal topic contents, we verified some common SA information for hurricane events such as stay safe advisories, death tolls, power outages, wind speed, donation and volunteer topics. In addition, we discovered specific SA information during Hurricane Laura, for instance, the warning that using a generator can cause carbon monoxide poisoning were posted on Aug 28, the same day when the FEMA provided the same advisory (Table 2.7); death toll numbers were posted on Aug 28, two days earlier than the official report from Louisiana Department of Health (Table 2.8); and an update of outage information in Lake Charles was posted on Aug 26, closely following the situational report of Department of Energy (Table 2.9).

The results of our work from section 4.1, 4.2 and 4.3 had twofold benefits. First, it provided a unique classification schema of specific events for hurricane and disaster related research. Second, it supplemented SA for practitioners who were affected during the disaster events, including federal and local agencies and public citizens.

From a research perspective, our study provided the numbers and keywords of latent topics during the Hurricane Laura. The state-of-the-art work that using tweets for retrieving SA to support disaster response with supervised models mostly assigned each tweet a single label [1,81]. Among these tweets, however, most actually contained information across multiple topics. For example, the first tweet in Table 2.7 contained both topics of "Advisory" and "Emotion"; the third tweet in Table 2.11 involved both topics of "Information Source", "Damage", and "Advisory". To fully utilize information and further inform precise situational awareness, our model could serve to generate reasonable classification schema and assign each tweets multiple labels in a supervised model.

From the practitioner perspective, the emerging topics captured by our model can assist the state and federal agencies in casualty and damage estimation with low-cost and wise resource allocation. For example, the agencies can prioritize tasks in different ways.

## 2.5   Conclusions

Retrieving real time SA information from big data of tweets is state-of-the-art methodology to aid disaster response and management. However, there remain several quantitative and qualitative challenges to understand and summarize first-hand tweets during disaster events.

One of the challenges to analyze tweets during new hurricane events with a supervised model is to first determine the classification schema into which the tweets should be classified. These categories should ideally be representative of the potential topics in the sense that they reflect the public concerns and issues caused by the event. To address this problem, we applied an LDA topic model on the prep-rocessed 10-day tweets collected during Hurricane Laura. Then human experts observed the automatically generated topics and manually assigned and organized category labels for them. Our tested LDA model with a Guided approach can make topics more humanly interpretable and determine classification schema for further analysis.

The other challenge is to rapidly and precisely update and extract SA information to government agencies and public citizens. Our study combines a sequence model and topic visualization to investigate the daily latent topics during Hurricane Laura, and evaluates the representative tweets for each topic by comparing with authoritative announcements and reports. The results suggest that agencies should adopt and adapt strategies day-by-day and wisely allocate resources based on affected areas and population because the number and content of SA topics varies across different stages of disaster events.

Nevertheless, there are still some potential limitations in our current work. First, due to the co-occurrence of several events including COVID-19, wildfires in California, he presidential election and other hurricane events, most tweets were hybrids, containing information cross events so that it is difficult to independently analyze the topics solely for Hurricane Laura. Second, our LDA model only reveals potential categories but assigns labels to each tweet data. In the future, it is worth apply-ing advanced attention model to experiment with multi-labels generated from our study and thus further inform situational awareness.

# Participatory Mobile App Development to Engage Citizen Science in Facilitating Environmental Pilot Studies

## 3.1 INTRODUCTION

Recently, many science disciplines demand and encourage broader engagement of not only scientists but also a variety of participants including government agencies, organizations and general public. Environmental science is one of these disciplines, and its route since last century has been outlined as "the three eras"[23]: the first era (1969-1992), environment information was created by specialists for specialists; the second era(1992-2012), it was produced still by specialists, but development of Web enabled its dissemination among public; the third era(2012-present), citizen science crowd sourcing data has been collected and integrated as supplement to official data.

Since the 19th century, citizen science has emerged to provide important observations for natural sciences[105]. The earliest definition of citizen science was how citizens accumulate knowledge in order to learn and respond to environment threats[56]. Compared to its historic form that only a privileged few could involve in the citizen science discipline, modern citizen science is potentially available to everyone who would like to contribute observations[7]. Thanks to the advancements in information technology, the social interconnection over the Web, and the widespread use of mobile devices, citizen science has consistently evolved along with the path of the three environmental eras.

Nowadays citizen science has been summarized as the scientific work that comes from the public either in collaboration or under the direction of professional scientists [22,89,27].

In a broad sense, the crowed sourcing data collected from citizen science may not always come with inherent geographic information. Nevertheless, spatial footprint and analysis are essential to most of environmental and geographic research. Under this circumstance, volunteered geographic information (VGI) was coined by Goodchild to refer the work where public volunteers contribute georeferenced distributed multimedia with the help of the diffusion of the Web and smart devices [41]. As citizen science and VGI have attracted more attention, more related projects and studies have been conducted in different disciplines everywhere over the world. In the domain of environmental science, citizen science and VGI play key role in tasks including data supplement, calibration and validation. Many researches targeting on monitoring of natural, environmental, human-driven and social changes and events have proved the rigour of crowdsourcing data and thus achieved success with low cost but high quality via citizen science and VGI [41,32,25,40]. At the same time, success has also landed in industry titans such as Google, Twitter, ESRI, OpenStreetMap, Wikeipedia and Sina Webo because they integrate citizen science and VGI into their business models.

The way that citizen science and VGI contribute to these successes can be classified into two categories: passive approach and active approach [73]. In passive approach, the data is collected without active involvement of citizen. For example, research can retrieve data(i.e. tweets) by using Application Programming Interfaces(API) provided by social media platform(i.e. Twitter). On the contrast, active approach refers that citizens voluntarily collect and send data, and this requires researchers and stakeholders to design and develop their own specific tools and applications. However, most hot topics related to citizen science and VGI solely focus on data analysis, quality and reliability while less discussions are opened up on the design and development of citizen science and VGI tools. In spite of the fact that data is the pivotal task in citizen science and VGI, it is important to understand and study the applications and platforms for data collection because some questions remain open. For instance, what are the characteristics and functionalities of different user interfaces for VGI creation on smart devices? How can stakeholders select relevant VGI for their specific tasks and needs? Therefore, we conducted two international studies, Survey123 in Poyang Lake and AppStuido for hurricane events to address these problems. This chapter introduces the two pilot projects, reports the experience, compares the findings, and discusses possible future developments, to offer extra case studies and provide recommendations on developing mobile applications for researchers and stakeholders.

## 3.2 RELATED WORK

The birth of OpenStreetMap (OSM—www.openstreetmap.org) has been considered as an important milestone in the VGI history[7]. Since OSM appeared more than ten years ago, new collaborative mapping approaches have emerged. To continue on the conceptual and practical exploration, Brovelli *et al.*[17] conducted a number of collaborative mapping experiments ("mapping parties") including five different dimensions and discussed the key outcomes and lessons learned from each of the mapping experiments. Mazumdar *et al.*[73] demonstrated the Crowd4Sat project that investigated different facets of how crowdsourcing and citizen science impact upon the validation, use and enhancement of Observations from Satellites products and services.Criscuolo *et al.*[23] reported on a test of using two VGI apps to collect either biological or abiotic observations along naturalistic trails, and discussed crucial characteristics of the app.

Although Twitter is the most popular social media platform to collect observations[76,87,98], its recent policy disables the geo-location acquisitions with free developer account, and it has a credibility issue regarding fake advertisements and robot tweets[72]. The National Park Service and U.S. Fish and Wildlife Service have encouraged volunteers to use eBird and iNaturalist to record observations of birds and other species in national parks and wildlife refuges[110,74]. However, these pro-grams requires expensive cost and professional programmers not only on the frontend in terms of user interface design and functional implements, but also on the backend for data storage, curation, maintenance and quality control services.

ESRI, the industry leader in GIS, has developed story maps to engage people with places (storymaps.arcgis.com). The story map templates allow non-specialists to link web maps with images and narrative and display the collection to public audiences[8]. With the help of the Esri Applications Prototype Lab (APL), Eanes *et al.*[30] created the "Wisconsin Geotools" that allowed communities to capture their own citizen observations that would accumulate into "spatial narratives". However, the story map tools relied on web application have limitations in terms of user experience and data collection in the field compared to mobile applications on smart devices. In this case, Survey123 and AppStudio were launched that allow users with less professional programming experience to build geo-enabled cross-platform native apps from a single code base according to different purposes. Since research and program using these two products have not fully discussed, we propose to use our pilot study to demonstrate the process to apply Survey123 and AppStudio in the development of mobile application to collect real word observations.

## 3.3 Application Development

In Chapter 1, we used Sentinel-1 imagery to retrieve the inundation area information of Poyang lake, but limited access to the DEM and water level data and lack of social economic status of local citizens impeded our further research on crane habitat conservation. In Chapter 2, we used Twitter as the data source to retrieve situational information. However, the challenges of limited query and disabled acquisition of geo-location via free Twitter REST APIs, low effective percentage of raw data, and version update constrained the efficiency and depth of research requiring spatial analysis. A solution is to develop in-house applications that satisfy all the needs we have for our studies.

In general, a development of web or mobile application consists of three tier system architecture (Figure 3.1): the presentation layer, the logic layer, and the data layer. The presentation tier is the front end layer providing the user interface. This user interface is often a graphical one accessible through a web browser or web-based application and which displays content and information useful to an end user. This tier is often built on web technologies such as HTML5, JavaScript, CSS, or through other popular web development frameworks, and communicates with others layers through API calls. The application tier contains the functional business logic which drives an application's core capabilities. It's often written in Java, .NET, C#, Python, C++, etc. The data tier consists of the database/data storage system and data access layer. Examples of such systems are MySQL, Oracle, PostgreSQL, Microsoft SQL Server, MongoDB, etc. Data is accessed by the application layer via API calls.

A simple example of a 3-tier architecture in our study, from user perspective, would be when volunteers collected field observations using our application. As end users, they might access a survey or a map through either the Survey123 or AppStudio interface which was the presentation tier. Once they interacted with widgets that information was passed on to the application tier which would query the data tier to call the information back up to the presentation tier. This happened every time when the volunteers submitted surveys or observations.

The additional advantages of a 3-tier architecture include agile development, scalability, and reliability. First, tier modularization enables the developer team to more efficiently enhance the application than developing a singular code base because a single layer can be upgraded with minimal impact on the other layers. Second, it is easy to scale each layer separately depending on the need at any given time. For example, if there are many submissions from a handful of volunteers, we can scale out the application and data layers to meet those requests without touching front end servers. Third, it can increase reliability because physically separating different parts of an application minimizes performance issues when a server goes down.

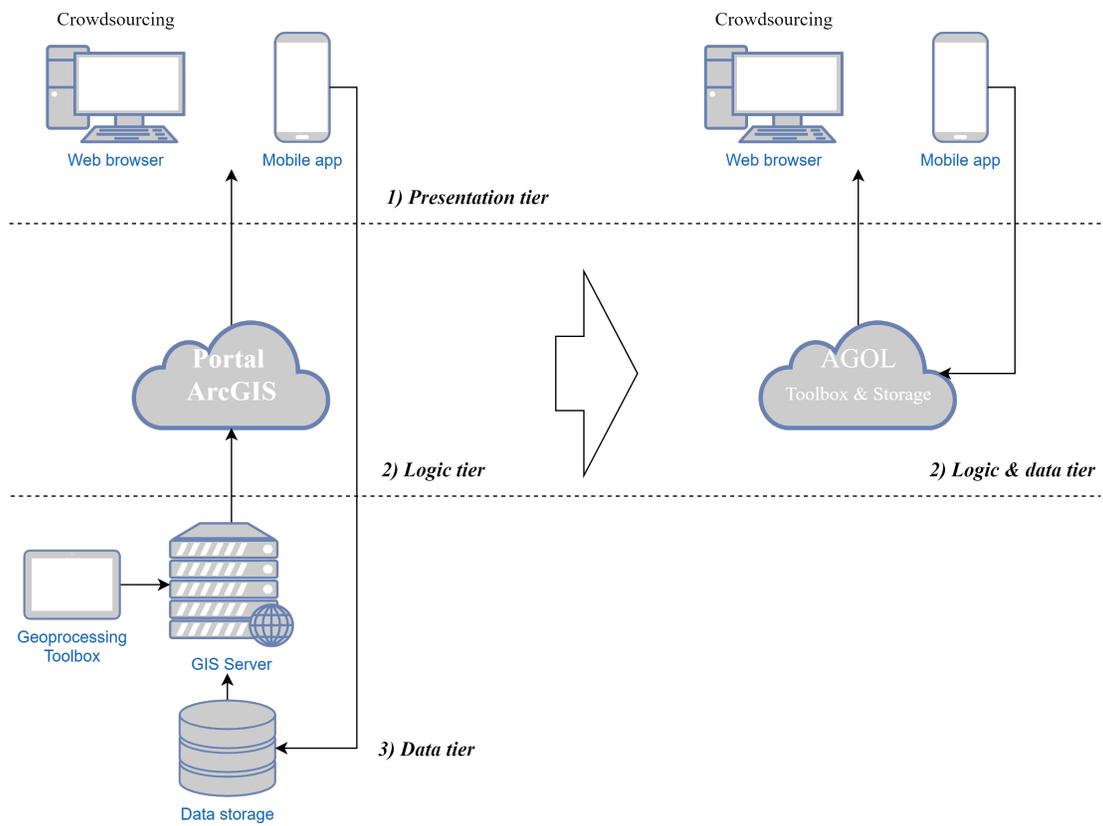

**Figure 3.1:** 3-tier architecture

Fully taking these advantages requires a developing team comprised of specialized developers in front-end, server back-end, and data back-end development and deployment. However, it is unrealistic to expect all research teams will have time, capital and technical skill to develop self-made platforms and applications[66]. Alternatively, a commercial and open-source platform (Figure 3.1) can help individuals initiate citizen science projects with two key features: a front-end user interface representing the presentation tier and a "sign up and go" back-end combining the logic tier and the data tier, such as Esri cloud SaaS platform- ArcGIS Online (AGOL), which ensures Webmap wizard, data storage and end-user friendly interface. Based on the AGOL system, Surver123 and AppStudio are two popular applications that have been used for citizen sciences. Survey123 is an easy-to-use mobile application that asks a series of questions to collect data and geospatial information from a smartphone or any other mobile device equipped with GPS, even when disconnected from the Internet. It can seamlessly integrate with GIS tools for further analysis. Similarly, AppStudio is a suite of productivity tools for building and deploying native apps to smartphones, tablets and laptop/desktop devices.

To remove the barriers and relax the constraints from previous chapters, we followed a specific evaluation metric of 3-tier architecture to choose suitable application for our studies.

### 3.3.1 PRESENTATION TIER

The presentation tier is the front-end interface where users can interact with different widgets that developers design. Thus, we selected metric from both user and developer perspectives. For user, the metric highlighted how easily user can input data with different devices (Table 3.1). For developer, the metric focused on how friendly a develop can learn and modify the set-up of front-end (Table 3.2).

**Table 3.1:** The metric for presentation tier from user perspective.

| Category | Description |
|---|---|
| Data input | The types of data that can be collected |
| Compatibility | Platform that supports the application |
| User interaction | The ability to see contribution from other users |
| Easy to use | The time and cost required to learning how to use |

**Table 3.2:** The metric for presentation tier from developer perspective.

| Category | Description |
|---|---|
| Development tool | The requirements of library and language to develop |
| Interface design | Availability of customizing widget options |
| Learning curve | The time and cost required to learning how to develop |

### 3.3.2 Logic Tier

Since users cannot access logic tier, the metric (Table 3.3) was only selected from developer perspective and it res presented how flexible the developer can customize the connection and use APIs.

**Table 3.3:** The metric for logic tier from developer perspective.

| Category | Description |
|---|---|
| Server | The online portal to support the application |
| APIs | The application programming interface to receive requests and respond |
| Customization | The ability to modify how the data from back-end transfer and display in front-end |

### 3.3.3 Data Tier

The data tier is hidden from view of users because of privacy and security concerns. The metric (Table 3.4) described how robust the built-in database is and the availability of various options.

**Table 3.4:** The metric for data tier from developer perspective.

| Category | Description |
|---|---|
| Database Support | The database supplied by the platform |
| Data Import | Options to import pre-existing datasets |
| Data Edit | The ability to edit the collected data |
| Data Export | The ability to export the collected data |
| Data Analysis | Specific tools that allow the author to geo-process |
| Data visualization | The availability of displaying data and providing unique base maps |

## 3.4 Experimental Studies

### 3.4.1 App Selection

The comparative analysis of using Survery123 and AppStudio to design, develop, and deploy the crowdsourcing platforms are summarized in Table 3.5, Table 3.6, Table 3.7 and Table 3.8. AppStudio is more flexible but more expensive to learn compared to Survey123 in both presentation and logic tier. Survey123 is more suitable to the target users who have less technology experience because its form-based concept while AppStudio is more suitable to the users who would like to see contributions from other users and complicated user interface. To continue the research in the Poyang Lake, our target users were fishermen who were expected to have low experience of digital devices and our needs required them to answer a list of prepared questions, so Survery123 was the perfect choice. To collect observations in different hurricane events, our needs expected more advanced options and the target users were expected to have rich experience in how to use mobile applications, so AppStudio could provide robust solution.

**Table 3.5:** Comparison for presentation tier from user perspective.

| Category | Survery123 | AppStudio |
|---|---|---|
| Data input | multiple medias | multiple medias |
| Compatibility | all platforms except Linux | all platforms |
| User interaction | No | Yes |
| Easy to use | Very easy | Support required |

**Table 3.6:** Comparison for presentation tier from developer perspective.

| Category | Survery123 | AppStudio |
|---|---|---|
| Development tool | None or Excel | Qt, JavaScript, ArcGIS Runtime APIs |
| Interface design | Limited | Flexible |
| Learning curve | Less time and cost | Need tutorial or coding experience |

**Table 3.7:** Comparison for logic tier from developer perspective.

| Category | Survery123 | AppStudio |
|---|---|---|
| Server | Invisible | Need tutorial or coding experience |
| APIs | Invisible | Need tutorial or coding experience |
| Customization | Invisible | Need tutorial or coding experience |

**Table 3.8:** Comparison for data tier from developer perspective.

| Category | Survery123 | AppStudio |
|---|---|---|
| Database Support | Yes | Yes |
| Data Import | Yes | Yes |
| Data Edit | Yes | Yes |
| Data Export | Less options | More options |
| Data Analysis | Yes | Yes |
| Data visualization | Yes | Yes |

### 3.4.2 SUVERY123 IN POYANG LAKE

In order to leverage local citizen information in our research, we piloted a survey of local fishermen to acquire sub-lake information at Poyang Lake, China in January 2018, which we presented at the US-IALE (International Association for Landscape Ecology) conference in Chicago in April 2018. This pilot study aimed to create a citizen-science mobile app to collect local water gate locations and elevations, and to capture fisherman choices in sub-lake water gate control and management and attitudes towards crane conservation.

We designed a mobile spatial survey using the Survey123 and interviewed fishers at 7 sub-lakes.This field survey included three regions: Poyang National Nature reserve, Nanjishan National Nature Reserve, and Kangshan Dam (Figure 3.2).

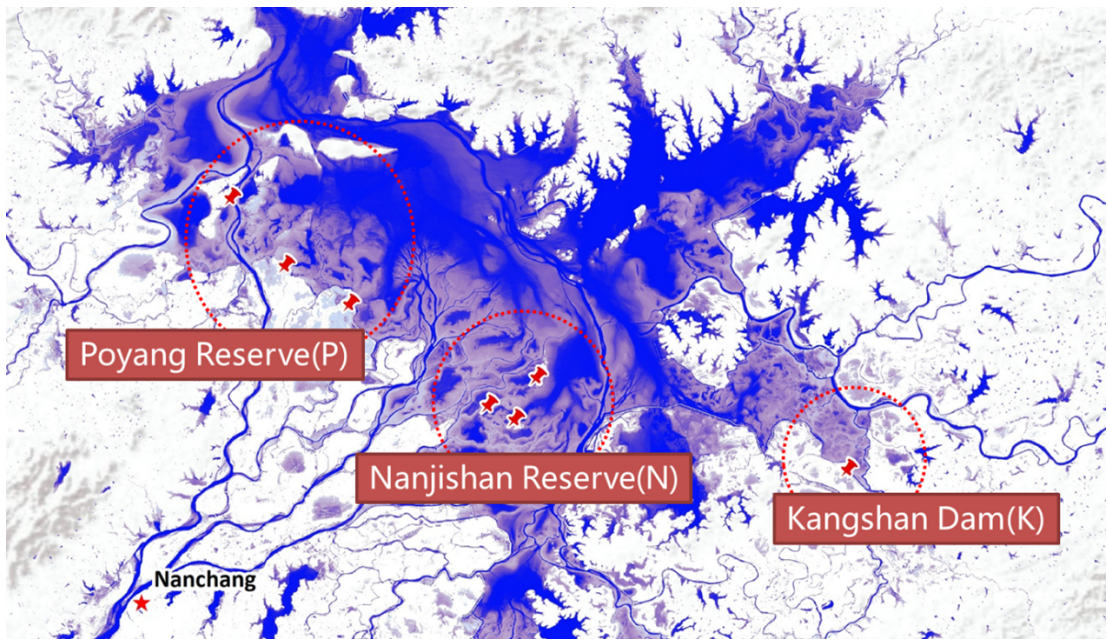

**Figure 3.2:** Winter pilot field study area with Survey 123

This spatial survery was organized into two parts with 16 designed questions on fishing, economics, hydrologics, and crane observations. The first part(Q1-Q14) were designed for the fishers while the second part(Q15, Q16) were designed for the recorders. The questions for each part were listed as follows:

Q1.  How big is your area of lake lease in mu (1 mu = 0.165 acre)?

Q2.  How much does the lease cost per year?

Q3.  How much do you earn per year?

Q4.  How many years have you been a lease holder?

Q5.  Which type of fish is your main economic source?

Q6.  Have you seen crane before?

Q7.  What types of cranes have you seen?

Q8.  Have you observed crane in your lease area?

Q9.  How do the numbers of cranes change according to your observation?

Q10.  Do you like crane? Why?

Q11.  Would you like to volunteer for monitoring cranes?

Q12.  Times of gate opening to impound water

Q13.  Times of gate opening to release water

Q14.  Times of gate closing

Q15.  Top of gate height (m)

Q16.  Water level this winter (cm)

These questions were prepared with XLSForm and converted to a spatial survey mobile app via Survey123 Connect. The final interface of the app (Figure 3.3) were comprised of clickable button, interactive map, and other input and uploading widgets.

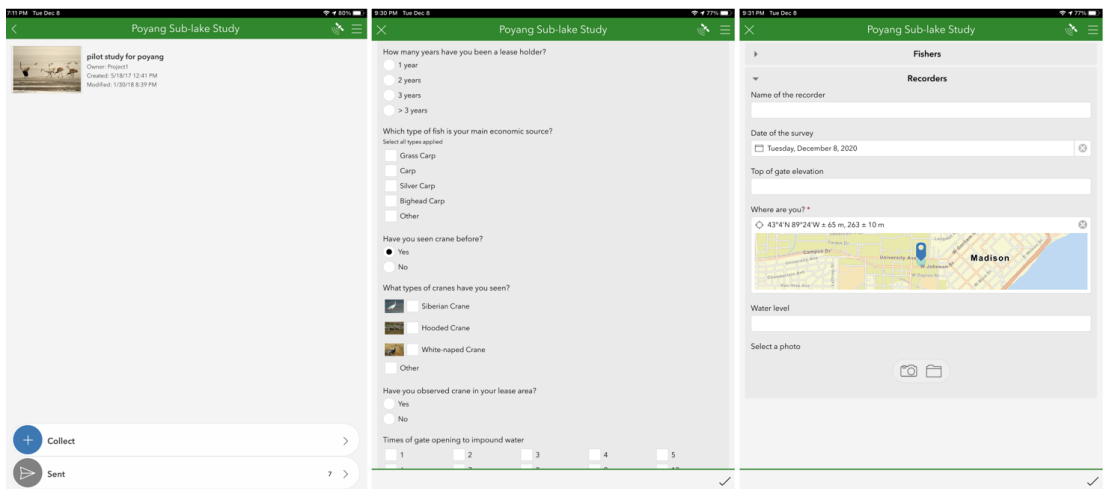

**Figure 3.3:** Spatial Survey Interface a. Login Interface b. Questions for Fisher c. Questions for recorders

Based on the responses from our economic questions, the sub-lake lease prize tends to be spatially related: the closer a sublake is to a reserve, the more expensive the lease price. The average cost in Poyang (P), Nanjishan (N) can reach $50K while only $17K in Kangshan (K) in the year of 2017.

From the feedback of crane observation questions, we could conclude that the local fishermen all have seen cranes wintering on their lakes. However, only common crane has been observed in all three regions. Siberian crane has been found in the two reserves, Hooded crane in Nanjishan, while White-naped crane was observed only in Poyang reserve. Unfortunately, these fishermen all responded with disliking cranes, because they believe cranes eat their fish, take oxygen from water, increase turbidity, and also make noises. However, they would like to volunteer for monitoring the cranes if there was a financial allowance.

The records from our hydrological questions reflected that the local fishery is using a weir system. Each wet season, some of the gates will be opened to impound water, and the fish will come with the water. Fishers then keep the gate closed to store water and grow fish. The gates will be opened again in dry season to release water to harvest the fish. These gates vary from 1.5 to 5 m in height, and the water depth ranges from $0 - 1.5$ centimeters. Also the year 2017 was abnormal because the Three Gorges Dam (TGD) opened twice on Oct to discharge flood waters which increased the water level again. We were also informed that the local government is trying to promote a policy called count birds reward lakes, basically, the government will send a team to each sub-lakes and subsidize fishermen by counting the birds. The total amount is estimated at $30K$ for 2017.

As a result, this survey has indeed verified the complexity of Poyang Lake system and demonstrated a strong economic contrast between fisheries and crane conservation. On one hand, the citizen level information indicated the seasonal fluctuation of Poyang Lake that further confirmed our results from the first chapter. On the other hand, the lack of local knowledge or appreciation for cranes impedes the cranes conservation.

### 3.4.3 AppStudio for Hurricane Events

This experimental study aimed to develop a participatory mobile app to harvest real-time geo-tagged observations during hurricane events. We designed our hurricane report application with AppStudio by customizing Quick Report template. The Quick Report template is a free-code solution for creating citizen engagement apps that allow users to capture an observation and submit it to an online service. As shown in Figure 3.4, the template has basic built-in functionalities to support selecting report type, adding geo-tagged observation and submitting report. However, it doesn't support advanced options, such as viewing all submitted observations, filtering observations, and displaying attached information of observations. These advanced functionalities are pivotal features for our requirements that users will be able to avoid submitting duplicate observations and share observations across the platforms. Thus, we customized the code for presentation tier and logic tier to add these advanced features.

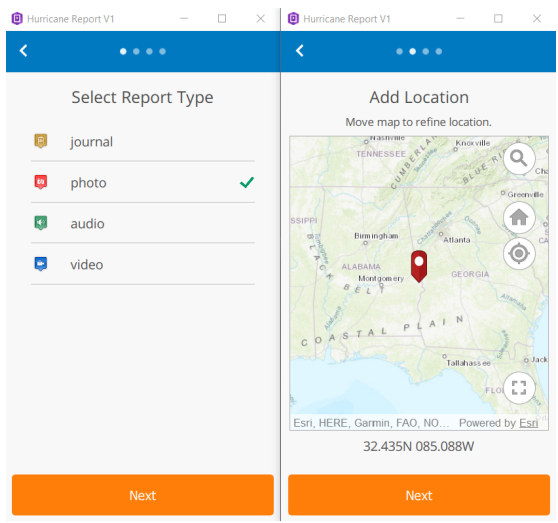

**Figure 3.4:** Quick Report Template

## AppStudio Design

The foundation of framework of AppStudio has three major components: Qt library, AppStudio AppFramework and ArcGIS Runtime for Qt. Qt Meta Lanuage (QML) is a JavaScript-based, declarative language for designing User Interface (UI) centric applications. The benefits of developing with QML including quick prototyping, component and code reuse, beautiful transitions and animations are results from its concepts of models, delegates and views. A model holds data consisting of a collection of items with the same type of properties, such as photos collected from observations. A delegate is a template that uses a common UI to render each model item on a view. A view displays the layout within which rendered model items are positioned. ArcGIS Runtime exposes the full capability of the ArcGIS platform to mobile, desktop, and embedded devices from simple map display and directions to advanced analysis and visualization. AppStudio AppFramework compliments and fills the gap for both Qt and ArcGIS Runtime, provides boilerplate functions and features, takes care of all cross-platform differences.

Supported by the framework in deep level, the structure of Quick Report template consists of 18

.qml files in pages folder (presentation tier) and more than 30 .qml files in controls folder (logic tier). The length of the code ranges from tens to thousands of lines (Figure 3.5). In addition to the original functionalities, we implemented new features to achieve our requirements. These additional features (Table 3.9) included: point features with customized symbol to show all the observations from different users, a dropdown menu to allow users to filter by type of observations, a reset button to restore default view, a popout window to display the attached attribute of the nearby observations when users clicked on a location.

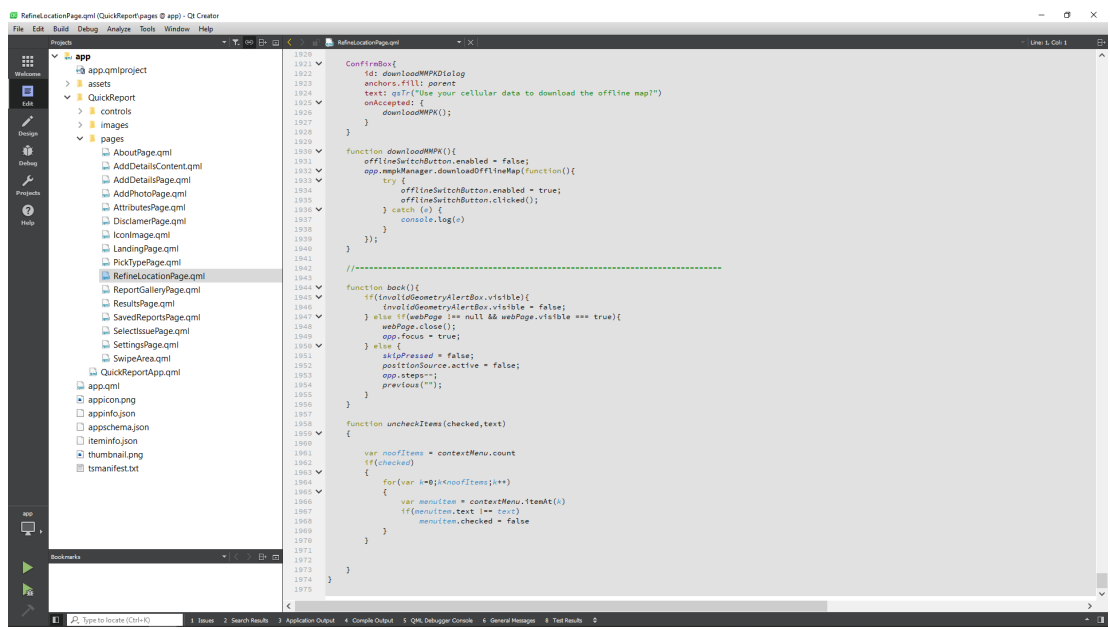

**Figure 3.5:** Qt IDE and sample code

The user interface of final version of application is shown in Figure 3.6, 3.7. Figure 3.6, from left to right, corresponds to open the application, select report types, display observations, and identify attached image respectively. Figure 3.7, from left to right, corresponds to filter observations, view attributes, upload media, and input details respectively.

We deployed and tested our application for Hurricane Sally and Hurricane Zeta during the 2020

**Table 3.9:** The new implemented features

| Feature | Implementation | Source file |
|---|---|---|
| Filter | ComboBox, ListModel, ListElement | RefineLocationPage.qml, ResultsPage.qml |
| Reset | Button | RefineLocationPage.qml |
| Popout | SwipeView, Repeater | RefineLocationPage.qml |
| Point | Pane, MapView, FeatureLayer | RefineLocationPage.qml |
| Attribute | dentifyLayerResult | RefineLocationPage.qml, EditControl.qml |
| Identification | onIdentifyLayerStatusChanged | RefineLocationPage.qml |

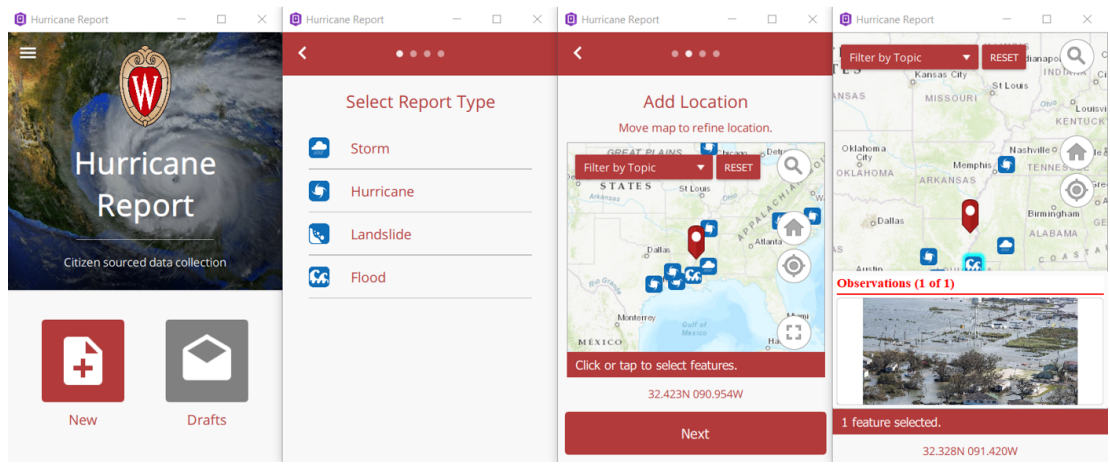

**Figure 3.6:** Hurricane Report App Final Version a. Login b. Event type c. Map view d. Attached Images

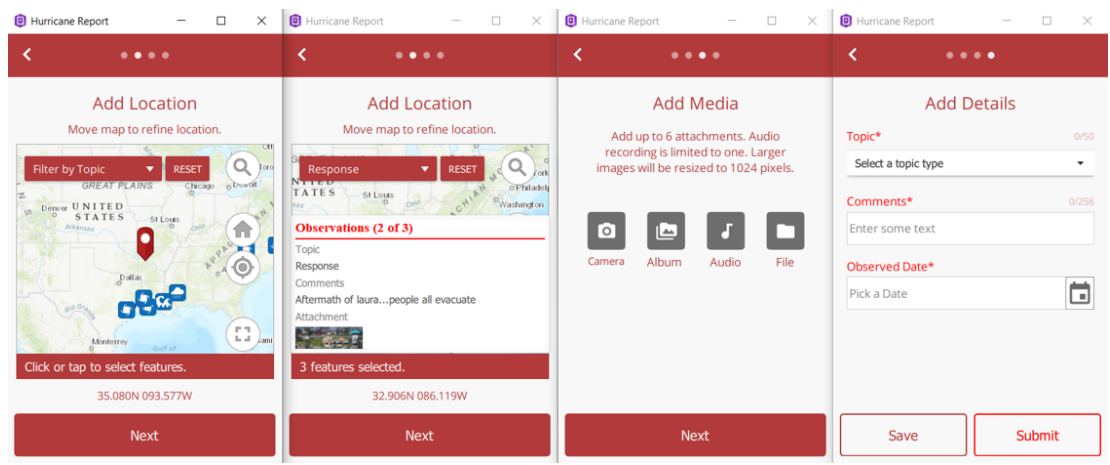

**Figure 3.7:** Hurricane Report App Final Version e. Filter f. Attribute g. Media type h. Submission

Atlantic hurricane season, which was the most active and the seventh costliest Atlantic hurricane season on record. The observations (Figure 3.8) were collected in the Courtland town on Virginia on September 20, 2020, and in the Elkin town across the Yadkin River on North Carolina on October 29, 2020. The photos depicts the flooding, road closure and emerged facilities with precise real time and geo-tagged location and attached short descriptions. Figure 3.9 shows the attached attributes of collected observations, the top red box corresponds to Hurricane Zeta while the bottom red box corresponds to Hurricane Sally.

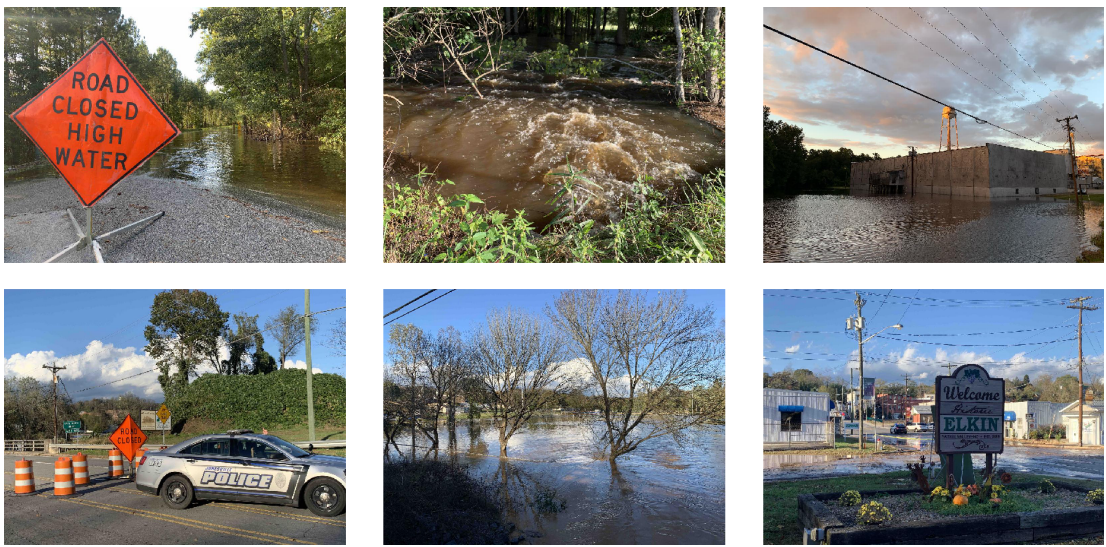

**Figure 3.8:** Observations collected during Hurricane Sally(Above) and Zeta(Below)

| Disaster Type | Topic | Comments | Observed Date | | Photos and Files |
|---|---|---|---|---|---|
| Hurricane | Impact | W Main and N Bridge St | 10/29/2020, 4:24:38 PM | | (1) Show |
| Hurricane | Impact | Closure 2 | 10/29/2020, 4:23:05 PM | | (2) Show |
| Hurricane | Impact | First responders | 10/29/2020, 4:20:39 PM | | (3) Show |
| Hurricane | Impact | Zeta | 10/29/2020, 4:17:16 PM | | (3) Show |
| Hurricane | Impact | USGS on site at gauge | 10/29/2020, 4:15:07 PM | | (1) Show |
| Hurricane | Impact | Gwyn McNeill Bridge, Hurricane Zeta | 10/29/2020, 4:13:27 PM | | (1) Show |
| Hurricane | Impact | Indian Town Rd and Meherrin Rd | 9/20/2020, 6:00:23 PM | | (3) Show |
| Hurricane | Impact | Peanut Factory underwater. | 9/20/2020, 5:55:53 PM | | (2) Show |
| Hurricane | Impact | Courtland VA Hurricane Sally | 9/20/2020, 4:33:51 PM | | (6) Show |

**Figure 3.9:** Attached attributes of observations collected during hurricane events

Compared to the tweets we scraped in Chapter 2, these observations contain more related information and less noises for retrieving situational awareness. More importantly, each comment has been tagged with predefined topic so that supervised machine learning models can be applied without extra costs for labeling task. In addition, the spatial information hidden in the Twitter can be acquired in the backend of my application which means that we can provide not only temporal but also spatial situational awareness in future hurricane events. Subsequently, the potential stakeholders can make wiser disaster response and management strategies in different days and regions. We can conclude that the primary test of our customized can successfully deliver important message to support disaster response and solve the constraints of using the Twitter as the crowdsourcing platforms.

## 3.5 Discussion

We presented two experimental studies to answer the question about characteristics and functionalities of different user interfaces for VGI creation on smart devices. The comparative analysis based on metric of 3-tier architecture from both user and developer perspectives and our workflow of developing and deploying Surver123 and AppStudio demonstrated what characteristics and functionalities were the prominence features in our study. As for the question on how stakeholders to select relevant VGI for their specific tasks and needs, our studies could also provide baseline examples to show the match up between the requirements of stakeholders and the functionalities of citizen science applications. Nevertheless, we would like to recommend to pay attention on some other factors of applications development beyond our studies.

### 3.5.1 Common Requirements of Stakeholders

It is difficult to enumerate all the requirements for all stakeholders because different stakeholders usually hold different requirements. From our own experience and previous research[47], we summarized the most common and basic requirements for different stakeholders into four different categories:

**Table 3.10:** Common requirements of different stakeholders regarding the app development

| Category | Requirements |
| --- | --- |
| Interface | Easy to access and use, self-explanatory, attractive design |
| Data contribution | Support in identifying features |
| Community | Opportunities for contact and exchange with other |
| Support | Guidance about how to collect and report data |

### 3.5.2 Version Control

Software versioning is a way to categorize the unique states of computer software as it is developed and released. Version control is a system that records changes to a file or set of files over time so that developer can recall specific versions later. Visioning issue is close related to the maintenance, management, upgrade, restore, and spread of the software applications. The state-of-the-art research on applications development for citizen science barely discussed this problem mostly because version control requires long-term tests. Even third-party, such as Qt library, usually controls version and makes sure the update will not affect across versions, it is unavoidable that some APIs, dependencies and libraries would be deprecated so that enable the applications nonfunctional or outdated. It is necessary to consider the version control when building applications.

### 3.5.3 Associated Costs of License and Maintenance

The associated costs can refer to any learning costs and incurred costs of operation and software downloads. Since the learning costs have been compared in previous section, the incurred costs of operation and software are remain open to discuss. Both Survey123 and AppStudio are not free and require an ESRI organizational account with publisher capabilities. Although they are among the most functional applications but they are also expensive options which means not everyone can afford the expenses. Alternatively, there are some completely open and free tools available but lack of detailed instructions and consistent functionalities. In this way, either option could potentially restrict the accessibility to citizen science projects.

### 3.6 Conclusions

Citizen science can break barriers and relax constraints in scientific research process with using the application development tools, specifically web and mobile participatory mapping platforms. The two ESRI software, Survery123 and AppStudio, are excellent options for stakeholders lacking the time and technical knowledge to create a data collection, management and visualization platform from the ground up. The comparison with metric of front-end and back-end, requirements of stakeholder, version control, and associated costs indicate that it is challenging to implement and harmonize the preferences, and to find a trade-off between technical feasibility, time and effort needed for application development and deployment. However, it is still necessary to understand the basic requirements of users and developers and be familiar with the functionalities of the technologies to create participatory applications that facilitate data collection for scientific research and leverage the importance of citizen science.

With demonstration of two international pilot studies on how to design and develop functional and suitable participatory mobile applications, our study is only a starting point that needs to be fur-ther refined. The development workflow is a circulating process need interactive feedback between users and developers. In other words, more citizens should be involved in the development process and, thus, delivering more suitable citizen science solutions for different stakeholders in the future.